\documentclass[5p,times]{elsarticle}

\usepackage{amssymb}
\usepackage{amsmath}
\usepackage{multirow}

\usepackage[hyphens]{url}
\usepackage{hyperref}
\hypersetup{breaklinks=true}
\usepackage{subcaption}

\journal{Nuclear Physics B}

\begin{document}

\begin{frontmatter}



\title{ A Two Degrees-of-Freedom Floor-Based Robot for Transfer and Rehabilitation Applications}



\author[label1]{F. St-Onge\corref{cor1}\fnref{fn1}}
\fntext[fn1]{This work was supported by the Fonds Québécois de la recherche sur la nature et les technologies (FRQNT) and the Natural Sciences and Engineering Research Council of Canada (NSERC).}
\cortext[cor1]{(Corresponding author: frederick.st-onge@usherbrooke.ca)}
\author[label1]{I. Lalonde}
\author[label1]{J. Denis}
\author[label1]{M. Lamy}
\author[label2]{C. Martin}
\author[label2]{K. Lebel}
\author[label1]{A. Girard}

\affiliation[label1]{organization={Department of Mechanical Engineering, Université de Sherbrooke},
            addressline={2500 Bd de l'Université}, 
            city={Sherbrooke},
            postcode={J1N 3C6}, 
            state={QC},
            country={Canada}}
            
\affiliation[label2]{organization={Department of Electrical Engineering, Université de Sherbrooke},
            addressline={2500 Bd de l'Université}, 
            city={Sherbrooke},
            postcode={J1N 3C6}, 
            state={QC},
            country={Canada}}

\begin{abstract}
The ability to accomplish a sit-to-stand (STS) motion is key to increase functional mobility and reduce rehospitalization risks. While raising aid (transfer) devices and partial bodyweight support (rehabilitation) devices exist, both are unable to adjust the STS training to different mobility levels. Therefore, We have developed an STS training device that allows various configurations of impedance and vertical/forward forces to adapt to many training needs while maintaining commercial raising aid transfer capabilities. Experiments with healthy adults (both men and women) of various heights and weights show that the device 1) has a low impact on the natural STS kinematics, 2) can provide precise weight unloading at the patient's center of mass and 3) can add a forward virtual spring to assist the transfer of the bodyweight to the feet for seat-off, at the start of the STS motion.
\end{abstract}








\end{frontmatter}



\section{INTRODUCTION}
\label{INTRO}
For patients in movement rehabilitation, accomplishing functional tasks is key to increasing quality of life and reducing the risk of rehospitalization~\cite{Functionalstatus,Motorandcognitive}. Training sit-to-stands (STS) is particularly useful as it has a significant correlation with increasing patient muscle power and the balance required to perform standing and walking tasks~\cite{BMAT2,STS_review}. Frequent training is essential to prevent muscle atrophy. However, studies indicate that up to 65\% of patients hospitalized in short-term care for seven days or longer develop muscle weakness due to prolonged immobility~\cite{ICUAW}. This is partly due to the current shortage of qualified clinical staff in hospital settings~\cite{Barriers_EM}. Clinical staff can use passive lifts to assist the patient's STS motion, such as the Guldmann GPT1 or the ARJO Sara Stedy, which hold the patient's and knees in a fixed position to reduce fall risks. However, using passive lifts for STS training can be exhaustive and lead to injuries for the clinical staff since it relies on them to move the patient's center of mass (CoM) \cite{Guldmann_GPT1,Sara_Stedy}. 

\begin{figure}
    \centering
    \includegraphics[width= 0.9\linewidth]{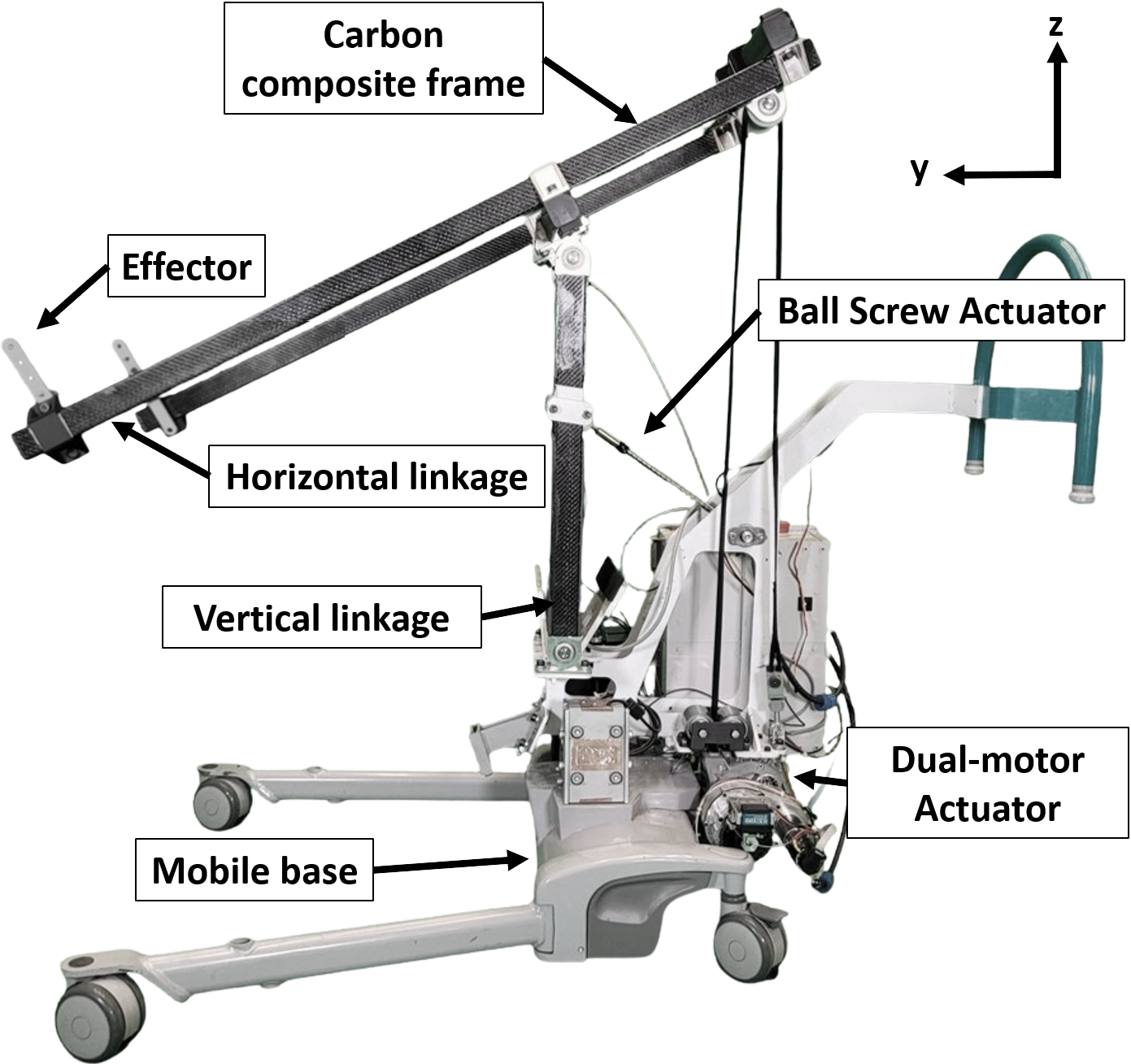}
    \caption{\footnotesize 2-DOF Sit-to-stand assistance robot}
    \label{proto_intro}
\end{figure}

Commonly found mobile robotic solutions in hospital settings include transfer raising aids, such as the Guldmann GLS5.2 Active Lifter~\cite{Guldmann}. These types of equipment can lift the patient's entire bodyweight on a fixed trajectory at low speeds, using a high force/low speed non-backdrivable actuation system. It is suited for transfer applications with low mobility patients who cannot participate in the STS motion, such as moving from the bed to the bathroom. It is, however, not optimal for rehabilitation training, as users need to be engaged in the motion to the maximum of their ability to train balance and increase muscle power~\cite{Changesinmusclepower,Failed_STS}. Furthermore, the fixed trajectory often does not adapt to the user's kinematics and prohibits the user from initiating the forward motion using the trunk swing.

Motorized devices dedicated to rehabilitation training provide higher speed assistance with backdrivable actuation to engage the users in the motion. Such solutions exist as ceiling-based lifts on rails, like the Aretech ZeroG~\cite{ZeroG_article}, but they are constrained to a rail and therefore require that the patients be transported to the dedicated "training room" each time, which is inefficient for frequent training to avoid muscle atrophy. Furthermore, for such products, the assistance relies on adjusting the tension and angle of the belt, meaning that the vertical and horizontal forces cannot be controlled independently, which 
limits the versatility for the development of new training strategies.

A multifunctional mobile device that combines the transfer raising capabilities (high-force/low-speed) and the rehabilitation training capabilities (low-force/high-speed) has not been developed yet. Such mobile device combination could be an opportunity: 1) for hospitals to save on dedicated equipment and rooms, 2) for the clinical staff to be more efficient by bringing the device directly to patients' rooms for training, and 3) for encouraging early patient rehabilitation training. This paper presents the design and testing of a multifunctional, mobile, floor-based sit-to-stand system as presented in figure~\ref{proto_intro}. The contributions are: 1) a robotic system that can alternate between a two DOF force-controlled configuration and a one DOF speed-controlled configuration, and 2) a force controller designed to allow different STS assisting modes, providing flexibility to the patient's level of mobility. The prototype provides four highly customizable modes. It can control the vertical (z-axis) and horizontal (y-axis) forces and impedances. It also has the transfer capabilities of commercially available raising aids. Lifting relies on a two-speed actuator to reduce the need for heavy and powerful motors. The prototype is tested with healthy adult participants. Section~\ref{state-of-the-art} first presents the state of the art. Section~\ref{Sys_req} presents the system requirements for each mode. Section~\ref{Des&Mod} presents the design of the robotic device and the controllers used for the various STS assistance modes. Section~\ref{experiments} presents the dataset acquired from experiments on healthy adult participants. Finally, Section~\ref{Disc&Concl} uses this dataset to verify that the requirements are met, discusses the limitations, and expands on future works.

\section{State of the art} \label{state-of-the-art}
Devices for STS training are of high interest in academic settings and can be divided into two broad categories: position-controlled systems and force-controlled systems.

Position-controlled systems assist the STS motion by guiding the user along a planned trajectory to simulate a typical motion, similar to transfer raising aids. They often take the form of a motorized platter or motorized handles that the users hold with their hands, sometimes with the addition of a harness to support the user's lower body~\cite{MOBOT,Skywalker}. For example, H.-G. Jun et al.~\cite{SMW} designed a system composed of a platter connected to a three-DOF mobile base. The platter, connected to the lower arms, also has an additional DOF which allows rotation for trunk inclination. Using EMG signals from the lower limbs and a 6-axis load cell in the platter, the researchers demonstrate that allowing trunk inclination reduces EMG signals and platter contact forces. B.~Sharma et al.~\cite{Platter_pHRI} expand on this idea by adding a force sensor-based admittance controller and an actuated rotating platter. This lets the user deviate from the planned trajectory and limits the human-robot interaction forces, with the benefit of allowing trunk rotation at the start of the STS. In both cases, the device partially supports the user's upper body, which is not the case during a typical STS motion. Also, imposing a trajectory or speed profile for STS assistance prevents the system from adapting to a user's kinematics. In summary, it does not let the user use its full muscle power to perform the STS, which can lead to atrophy for key STS muscle groups~\cite{Force_assistance_system,3DOF_Walker}. Controlling the forces applied is more representative of muscle activation and is more flexible to the user's kinematics.

Force-controlled systems assist the STS motion by directly applying forces to the user's body to reduce muscle load~\cite{Force_assistance_system}. D. Chugo et al.~\cite{3DOF_Walker} present an innovative three-DOF rehabilitation walker that can alternate between position control and force control. During STS assistance, the effector moves the user's upper body along a planned trajectory until it reaches a force threshold. R. Kamnik et al.~\cite{PDRAMA} designed a system that applies forces directly to the hips of the user by means of a motorized bicycle seat that can rotate. This time, the seat can apply independent forward and vertical forces, which are calculated using the contact forces read from load cells at the seat and instrumented handles to maintain upper body balance. Matjačić et al.~\cite{STS_trainer} designed a system that adapts the STS training according to a user's mobility level by changing the amplitude of the speed profile and the maximum output force provided to the lower body, obtaining assistive capabilities of up to 150~kg. However, restraining the lower body to the device does not let users train to balance their CoM, which is a critical skill to accomplish a successful STS~\cite{Momentum}. Furthermore, the transfer capabilities cannot match the payload requirement of commercial raising aid transfer devices, typically around 200~kg~\cite{Guldmann}. In summary, assisting the STS motion for force control is suited for rehabilitation training if the applied force reproduces the natural muscle activation of the body. 

Both position-controlled and force-controlled systems have their advantages for STS training and transfer, depending on the user's mobility level. However, they cannot cover the whole range of user needs, from high-force/low-speed transfer to low-force/high-speed partial bodyweight support. The next section describes mobility levels and how they are broken down into specifications for a multifunctional device.


\section{Assisting modes features}
\label{Sys_req}
Patient needs can vary widely, ranging from minimal assistance, such as a safety net for potential falls, to comprehensive support where the patient's whole bodyweight must be supported and guided along a path to maintain balance. The assistance provided by the device needs to be customizable to help as many patients as possible. To simplify this problem, four mobility levels (from A to D) are defined as examples of "typical patients" that could be encountered in a clinical setting, which were inspired by articles describing common STS failure causes, such as "step" and "sit-back" failures~\cite{BMAT2,Momentum,Failed_STS}. Each mobility level is then associated to an assisting mode designed to suit their training needs. These profiles are described below, in decreasing order of mobility.

\subsection{Mobility Level A: Follow Me Mode}
Patients with a mobility level A can perform a STS without assistance. The goal for clinical staff is therefore to test their muscle power and their balance during multiple repetitions. Since they can still be at high risk of falling due to various reasons, such as fatigue~\cite{BMAT2}, being connected to an assisting robot can limit the risks of injury. The Follow Me mode meets these requirements by acting as a safety net. No assistance is provided within the workspace to encourage these patients to use their muscle power and balance while following their motion to catch them in case of a fall. The design and validation of the fall recovery algorithm are not covered in this paper, but a working algorithm has been developed for a multifunctional ceiling-based robot with a similar actuation design~\cite{LM2S}. To allow these patients to move freely, the device needs to be as transparent to the users as possible. All resistance, including friction forces and robot inertia, must be minimized to avoid impacting key STS metrics. These key metrics in both the y and z axes are chosen as the total displacement, the peak velocity~\cite{STS_Traj_model}, the peak acceleration~\cite{Acceleration} of the CoM, and the peak ground reaction forces~\cite{STS_Event_std}.

\subsection{Mobility Level B: Weight Unloading Mode}
Patients with a mobility level B can generate the forward CoM momentum necessary for seat-off and then maintain the balance of their CoM while rising. However, these patients lack the full lower limb muscle power critical for lifting their CoM after seat-off and accomplishing the STS motion~\cite{minimum_muscle}. The goal for clinical staff is to compensate for this muscle weakness with a vertical force while allowing them to train their balance. As these patients improve, the vertical assistance is lowered to encourage them to exert more muscular force.~\cite{BMAT2} The Weight Unloading mode meets these requirements by applying a constant z-axis force all over the workspace and without kinematic restrictions on the user, allowing these patients to train their balance. The output vertical assistance at the effector must be within 5\% of the desired value during the whole motion to remain within the human weight discrimination capability at the CoM~\cite{Weight_dis2}. STS trajectory and speed profiles in the forward and upward directions of the user should therefore not be affected.

\subsection{Mobility Level C: CoM Balance Mode}
\label{C_mode}
Patients with a mobility level C need further assistance to perform the STS motion. Besides lacking muscle power to lift their CoM upwards, they lack the critical ability to generate forward momentum to achieve seat-off and therefore avoid "sitback" failure~\cite{Momentum,Failed_STS}. To assist these patients, the clinical staff provides both forward and upward forces to help the STS motion. The CoM Balance Mode generates a constant vertical assistance force and a forward force that reduces after seat-off to assist the patient's forward momentum only at the start of STS.

The CoM Balance mode meets these requirements by adding a virtual spring to vary the y-axis (forward) force with STS progression while keeping the same vertical assistance force as the Weight Unloading mode.
As the patient becomes increasingly skillful in generating forward momentum and maintaining their balance, the overall assistance can be lowered by reducing the rigidity of the virtual spring to encourage the patient to participate more in the STS motion.

\subsection{Mobility Level D: Transfer Mode}
Patients with a mobility level D cannot participate in the STS motion due to very low muscle power and balance. However, they still need to be verticalized from a sitting position for their personal needs, for example, to go to the bathroom. The system therefore needs to perform a transfer, where the device supports entirely the motion and imposes the CoM trajectory. The Transfer mode meets these requirements by recreating the simple spline effector trajectory commonly found on one-DOF commercial transfer raising aids, such as the Guldmann GLS5.2 Active Lifter~\cite{Guldmann}. To do so, the effector must be able to move users up to 205~kg (450~lbs) along the trajectory (see the black dashed line in figure~\ref{Ctrl_modes} e).

\subsection{Requirements Summary}

These modes aim to demonstrate the versatility of the proposed device’s assistance. This flexibility would allow clinicians to tailor rehabilitation plans more effectively to each patient's needs. Each mode should therefore be obtainable by modifying the same set of parameters, which are the desired weight unloading expressed as a percentage of the user’s bodyweight ($F_{z}$), and the desired rigidity of the y-axis virtual spring ($K_{y}$). 
The user’s height and weight are other inputs of the system: the height for estimating the y-axis (horizontal) standing position and the weight for converting the desired \%  bodyweight assistance into a target force. Table~\ref{param_modes} summarizes the desired parameter values for each assisting mode.

\begin{table}
\centering
\caption{Adjustable parameter values for each assisting mode}
\begin{tabular}{ccccc}
\cline{1-5}
                                 & A                          & B                          & C                          & D                          \\ \hline
\multicolumn{1}{c}{Controller} & \multicolumn{1}{c}{Force} & \multicolumn{1}{l}{Force} & \multicolumn{1}{l}{Force} & \multicolumn{1}{l}{Speed} \\ 
\multicolumn{1}{c}{$F_{z}$ (\%bodyweight)}  & 0 (safety net)                          & \textgreater{}0            & \textgreater{}0            & N/A                        \\
\multicolumn{1}{c}{$K_{y}$}         & 0                          & 0                          & \textgreater{}0            & N/A                        \\ \hline
\end{tabular}
\label{param_modes}
\end{table}

Note that the CoM Balance is the only mode that uses the $K_{y}$ virtual spring to help with forward momentum at the start of the STS motion.  The y-axis virtual spring approach lets the y-force decrease with STS progression, as explained in section~\ref{C_mode}. The virtual spring's anchor point is along a vertical axis located at the estimated final y-axis position of the CoM, calculated using equation~\ref{Ky_axis_pos}.

\begin{equation}
 \label{Ky_axis_pos}
    Y_{K_{y}\text{axis}}=E_{yi} + 0.25(\text{Height})
\end{equation}

\begin{figure}
    \centering
    \begin{subfigure}[b]{0.4\linewidth}
        \centering
        \includegraphics[width=\linewidth]{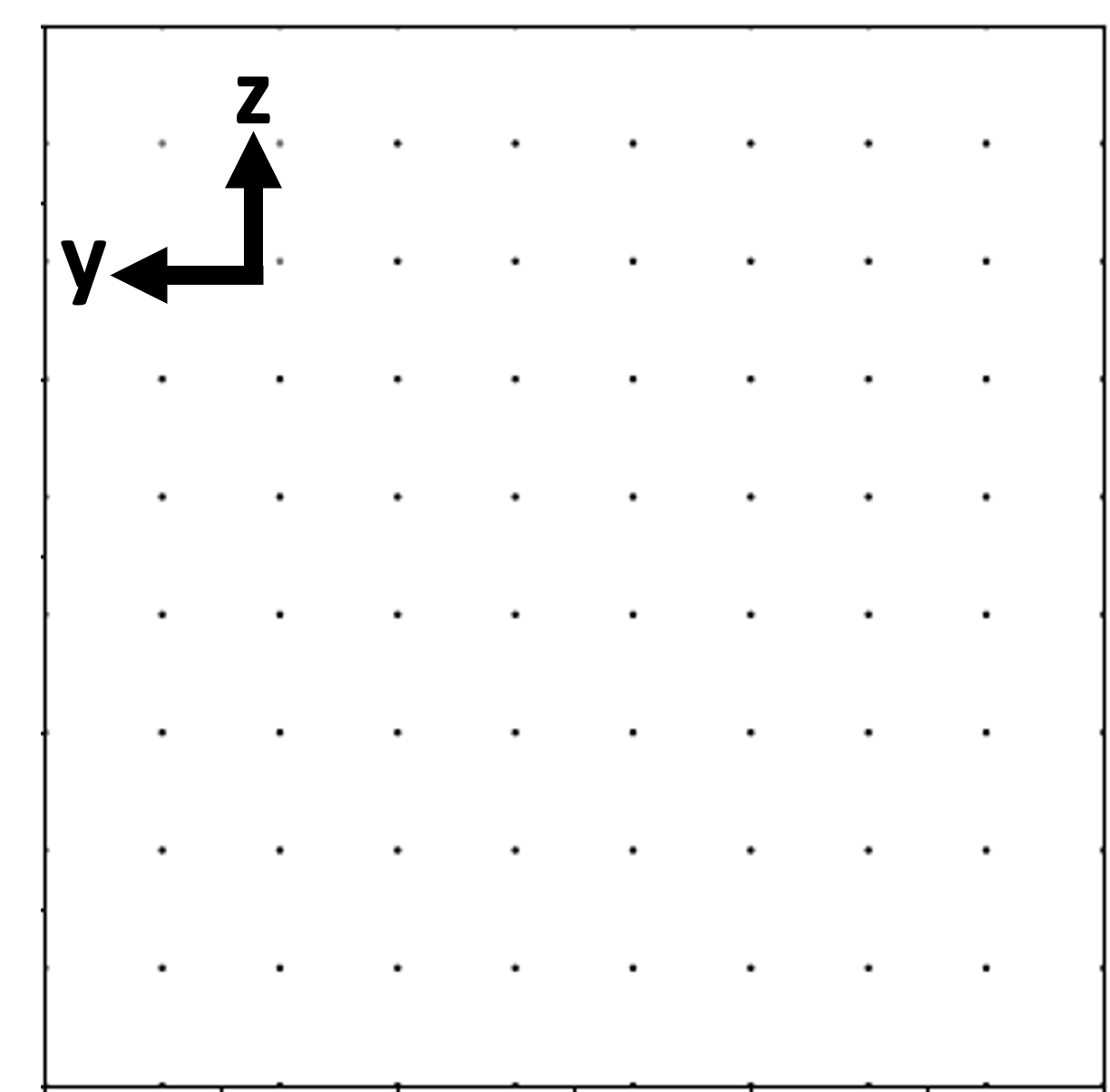}
        \caption{A mode}
    \end{subfigure}
    \begin{subfigure}[b]{0.4\linewidth}
        \centering
        \includegraphics[width=\linewidth]{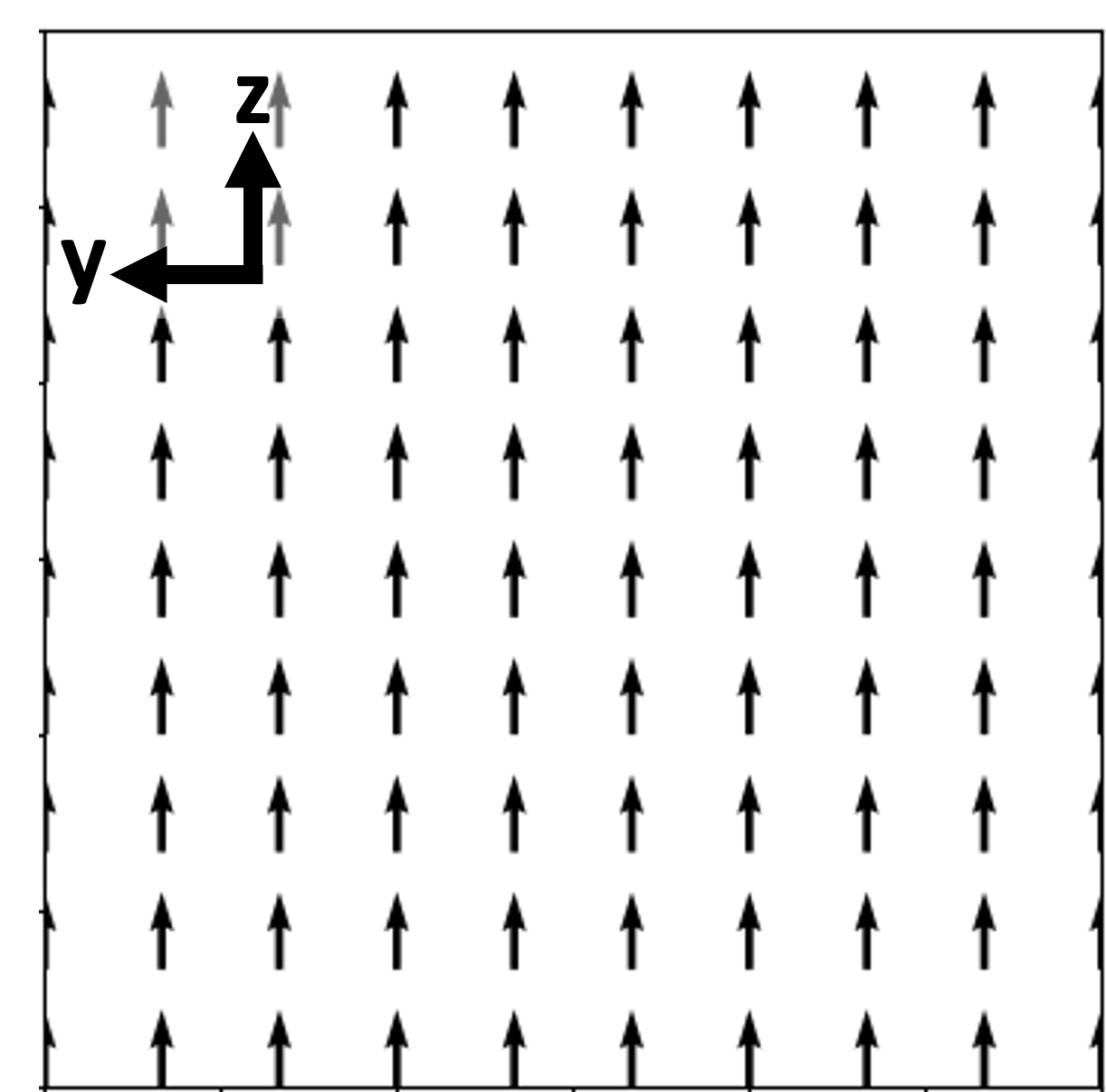}
        \caption{B mode}
    \end{subfigure} \\
    \begin{subfigure}[b]{0.4\linewidth}
        \centering
        \includegraphics[width=\linewidth]{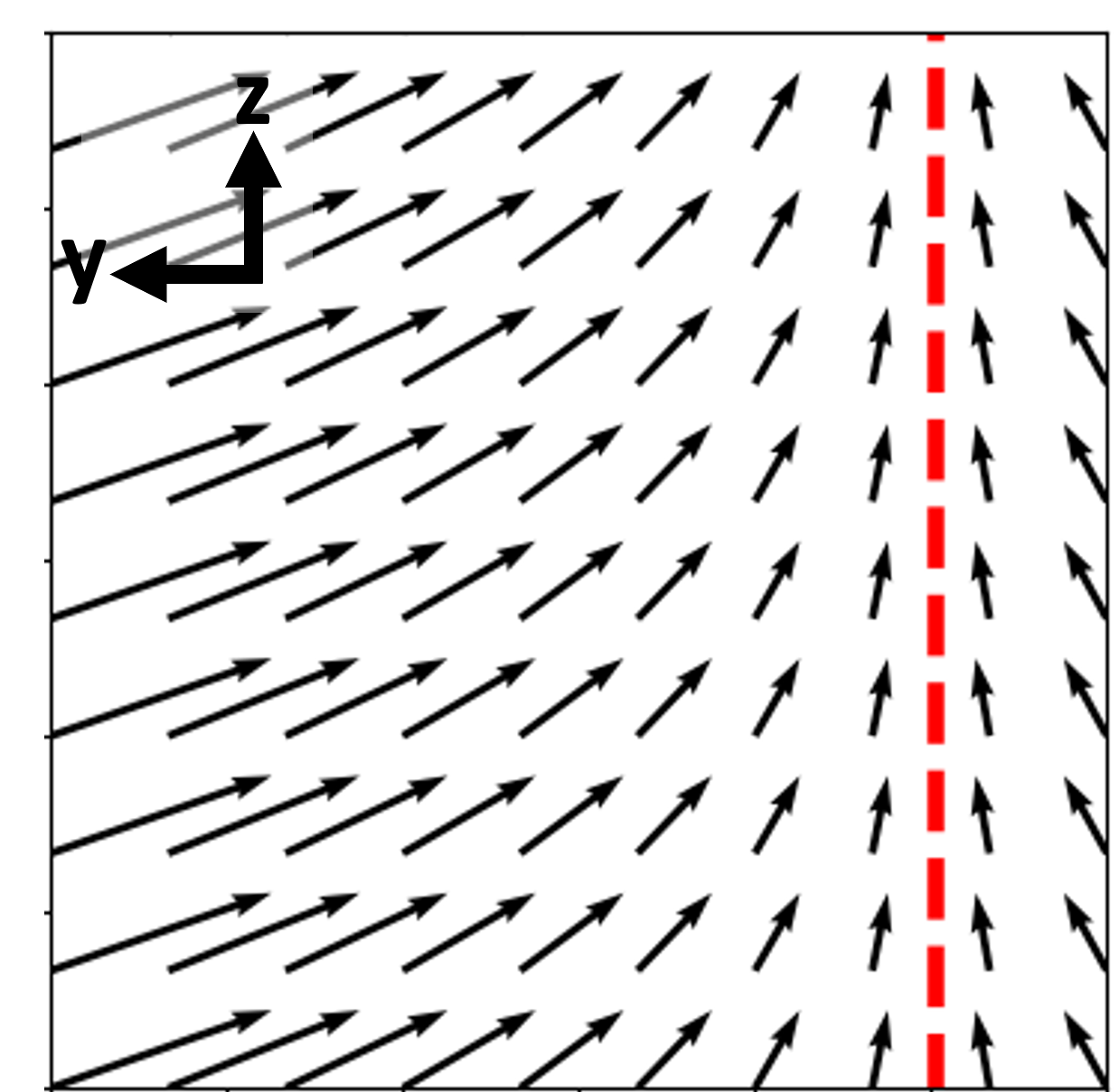}
        \caption{C mode}
    \end{subfigure}
    \begin{subfigure}[b]{0.4\linewidth}
        \centering
        \includegraphics[width=\linewidth]{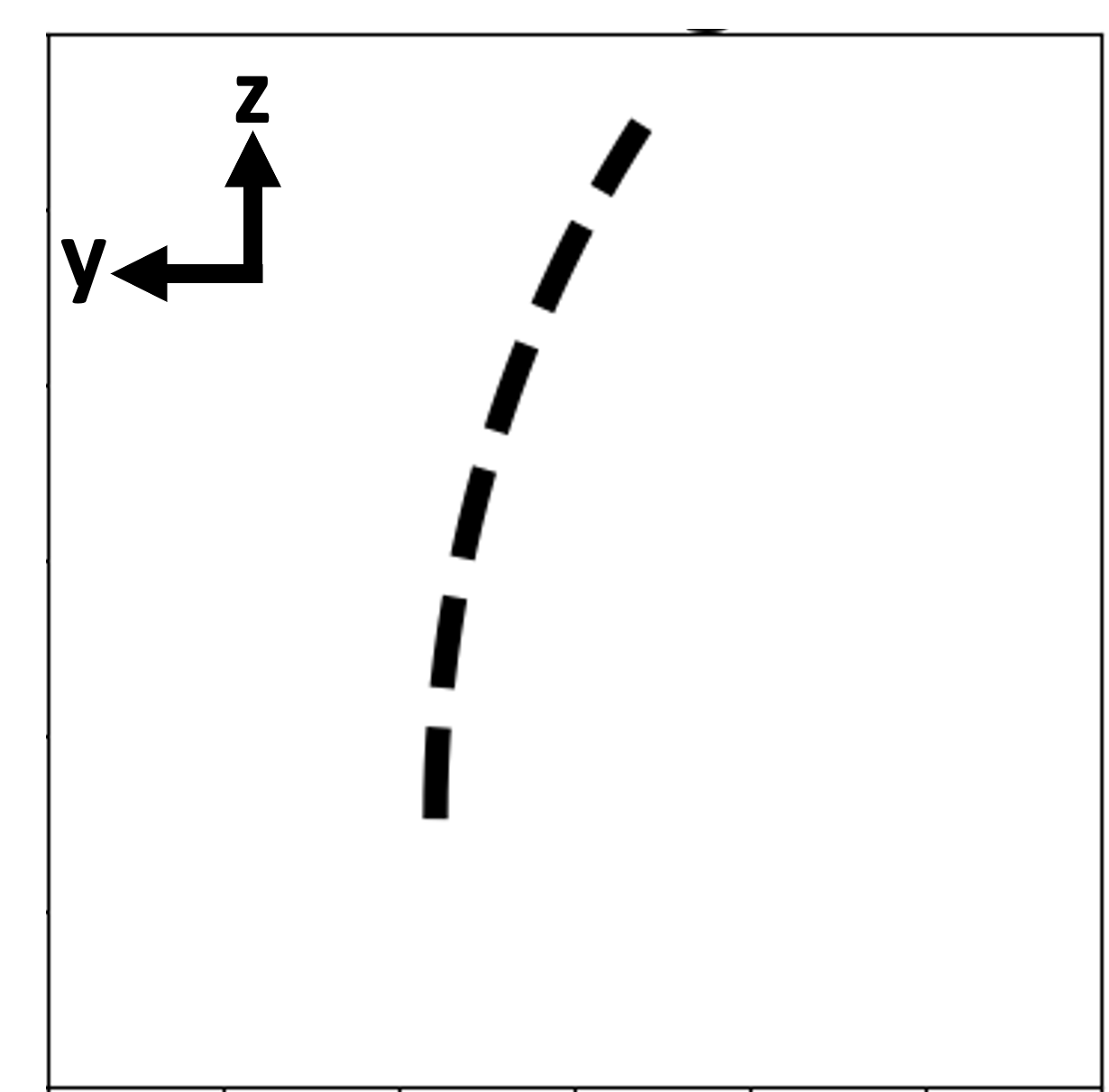}
        \caption{D mode}
    \end{subfigure}
    \caption{\footnotesize Workspace effector force field for control modes. For modes A to C, the black arrows represent the orientation and magnitude of the assistance force from the force controller deployed to the patient's CoM according to its position. For C mode, the red vertical axis represents the anchor point for the virtual spring $K_{y}$. For D mode, the dashed line represents the fixed trajectory from the speed controller.}
    \label{Ctrl_modes}
\end{figure}

Where $Y_{K_{y}\text{axis}}$ is the y-axis position of the vertical axis, $E_{yi}$ is the y-axis position of the effector at the start of the STS motion and Height is taken from the inputs of the controller. The addition of 25\% of the user's height to the starting y-axis position corresponds to the average ratio of the thigh length to stature~\cite{Percentile}. Also note that, as shown in table~\ref{param_modes}, the adjustable parameters are not used in Transfer mode as it uses a speed controller instead of the force controller used for modes Follow Me and CoM Balance. More details on this specific aspect will be provided in section~\ref{Des&Mod}. Figure~\ref{Ctrl_modes} illustrates the vector field of assistance for all assisting modes (A--C) and the imposed trajectory of the Transfer mode D.
%
Table~\ref{Sum_req} summarizes the output requirements at the effector for the "Rehabilitation" force controller and the "Transfer" speed controller.

\begin{table}
\centering
\caption{Effector output requirements}
\begin{tabular}{ccccc}
\cline{1-5}
       & \multicolumn{2}{c}{Rehabilitation}        & \multicolumn{2}{c}{Transfer} \\ \hline
Axis & $\textit{F}_{max}$ (N) & $V$ (m/s) & $\textit{F}_{max}$ (N) & $V$ (m/s) \\ \hline
Y    & 200      & 0.33~\cite{STS_CoM_vel}    & N/A      & N/A     \\
Z    & 650     & 0.35~\cite{STS_CoM_vel}    & 1962~(200~kg)     & 0.03    \\ \hline
\end{tabular}
\label{Sum_req}
\end{table}

For rehabilitation, the velocity requirements are based on maximum CoM velocities in healthy adults from clinical data on the STS motion~\cite{STS_CoM_vel}. In the y-axis, the maximum force value represents the peak fore-aft GRF (10\% bodyweight) deployed during an STS, which for the maximum lifting capacity of 200~kg in Transfer mode, represents 200~N~\cite{STS_Event_std}.

\section{Robotic System Concept} \label{Des&Mod}
This section details the proposed multifunctional STS assistance robot. First, the design of the structure is presented with its advantages to assist both rehabilitation training and transfer tasks. Then, the kinematics equations of the robot arm and actuators are derived to obtain the velocity and force requirements for both actuators, given the requirements at the effector output from the previous section. The actuator designs are then presented. Finally, the force controller for Follow Me to CoM Balance modes (including the structure's mass and actuators' friction compensations) and the speed controller for the Transfer mode are detailed.

\subsection{Architecture Design}
\begin{figure}
    \centering
    \includegraphics[width= 0.7\linewidth]{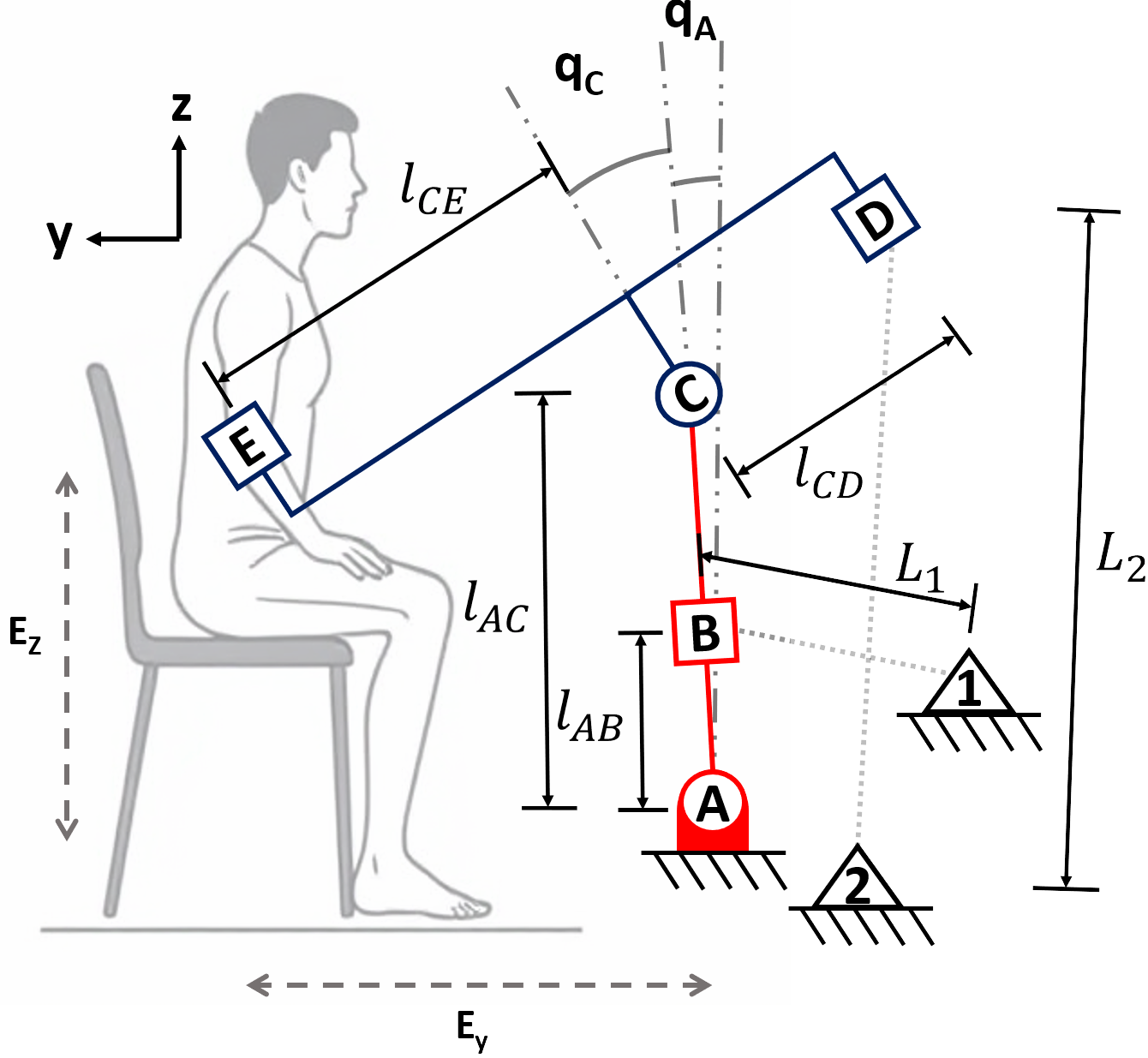}
    \caption{\footnotesize Model of the joints, linkages and actuators of the robot's arm. The structure is divided into two main segments, the horizontal linkage (linkage CDE) in blue and the vertical linkage (linkage AC) in red. Angles $q_{A}$ and $q_{C}$} represent the respective angular positions of each robot joint. $l_ab$ = 0.38~m, $l_ac$ = 0.61~m, $l_ce$ = 0.75~m, $l_cd$ = 0.38~m.
    \label{fig:model}
\end{figure}

The robot is designed as a two degrees of freedom (DOF) planar robot arm on a mobile platform as shown in Figure~\ref{proto_intro}. This choice provides independent y and z-axis forces in the user's sagittal plane while following various patient kinematics. Horizontal (y-axis) forces help the patient generate forward momentum and maintain balance during the STS, while positive vertical (z-axis) forces help the patient in lifting their CoM upwards. Figure~\ref{fig:model} presents the model of the robot's arm, where 1 and 2 are the actuators, A and C are the joints of the frame, B and D are the connection points of actuators 1 and 2 respectively, where their output force is applied on the structure. Furthermore, $L_1$ and $L_2$ represent the distance between each actuator and their respective connection points on the structure. Finally, Joint E represents the robot end effector, which is the connection point of the custom harness.

The first DOF is the rotation of linkage AC at joint A relative to the floor-lift base frame ($q_A$), and the second DOF is the rotation of linkage CDE at joint C relative to linkage AC ($q_C$). The user connects to the robot arm using a custom harness with a swivel connection at joint E for free rotation of the trunk. Joint A is 0.44~m above ground due to the height of the mobile base.

\begin{figure}[h]
    \centering
    \begin{subfigure}[b]{0.8\linewidth}
        \centering
        \includegraphics[width=1.0\linewidth]{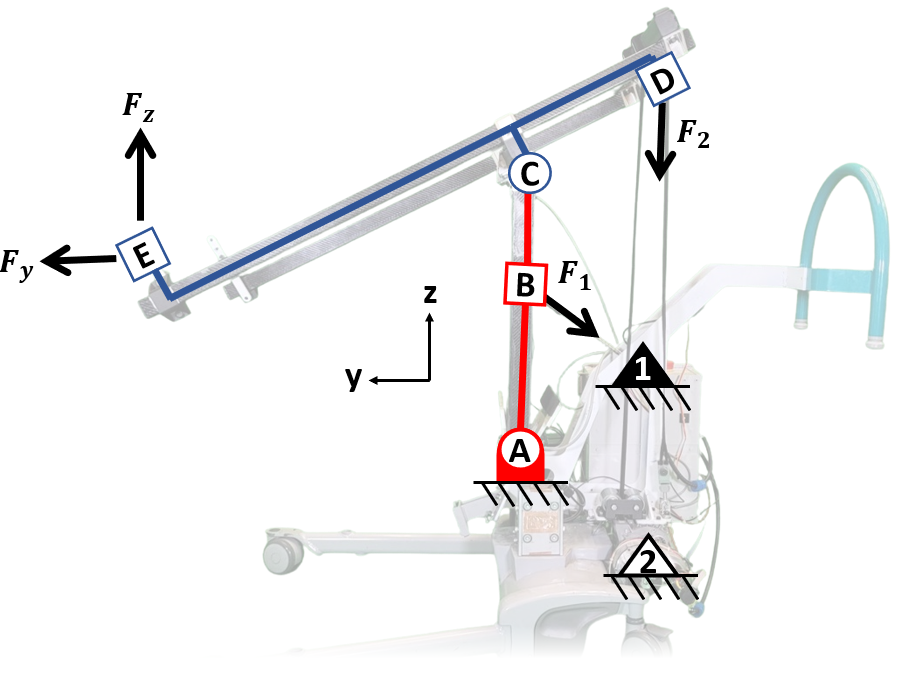}
        \caption{\footnotesize Force-controlled configuration of the device (Follow Me to CoM Balance modes)}
    \end{subfigure}
    \\
    \begin{subfigure}[b]{0.8\linewidth}
        \centering
        \includegraphics[width=1.0\linewidth]{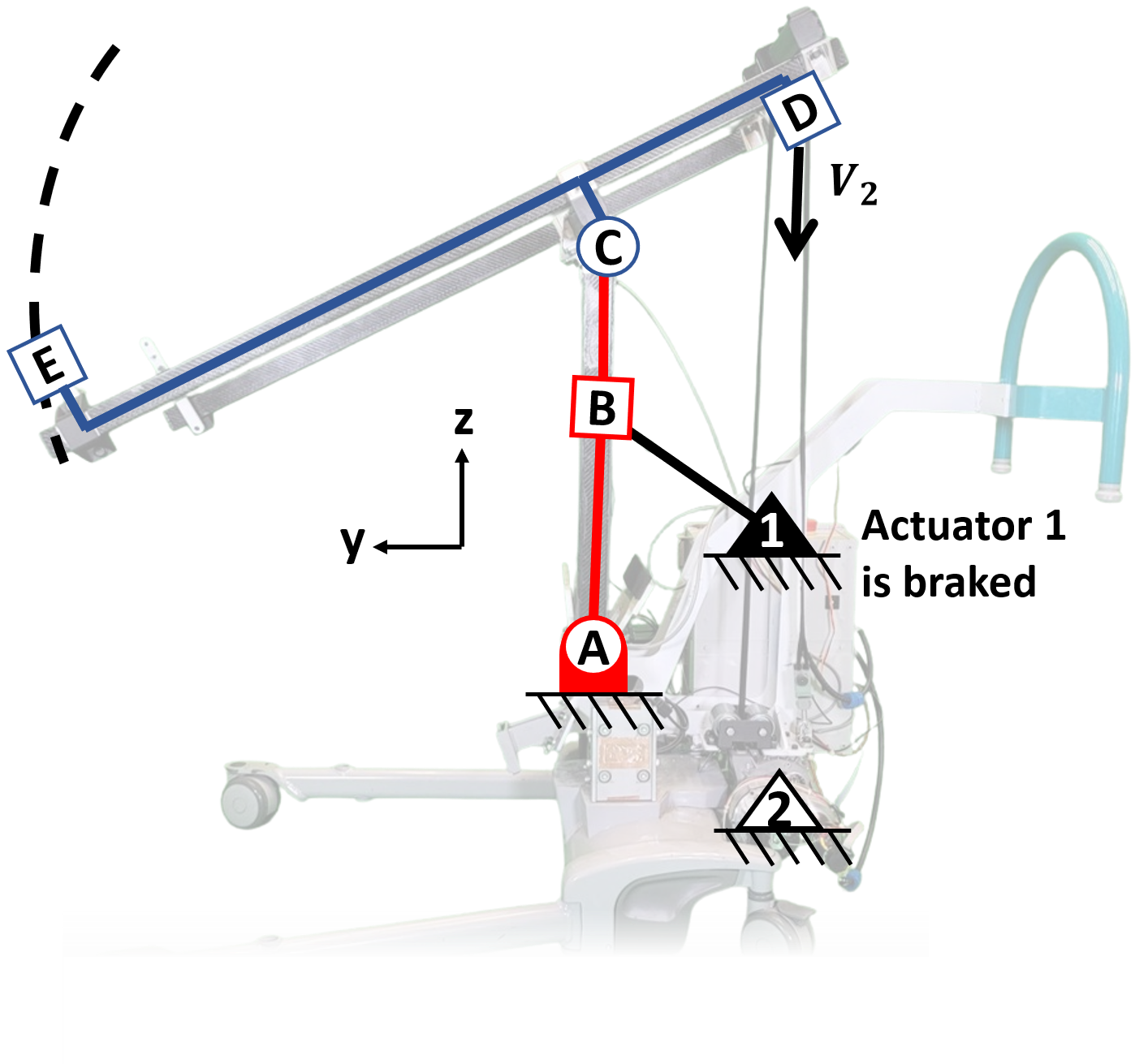}
        \caption{\footnotesize Speed-controlled configuration of the device (Transfer mode)}
    \end{subfigure} \\
    \caption{\footnotesize Configurations for force-controlled and speed-controlled of the device. For force control, both DOFs are free and both actuators can apply forces on the structure. For speed control, the device is restricted to one DOF around joint C, and only actuator 2 can apply forces on the structure.}
    \label{Configurations}
\end{figure}

To improve force control for Follow Me to CoM Balance modes, the inertia of the linkages is minimized using aluminum joints and carbon fiber square tubes and plates. The total weight of linkage AC is 2.65~kg and 4.91~kg for linkage CDE. Also, the actuators are not located on the linkages to further minimize inertia. Transfer Mode only uses joint C to emulate the one-DOF trajectory of commercially available raising aid devices. Actuator 1 uses a disc brake in this mode for holding the vertical linkage AC in place and thus does not need a powerful motor that meets all requirements. With this design, only the motorization of actuator~2 needs to bear the loads of the Transfer mode forces specifications, as shown in figure~\ref{Configurations}. The detailed design of both actuators 1 and 2 is presented in section \ref{Act_des}.

\subsection{Kinematics Equations}
The robot kinematics are derived from equations~\ref{raf} to~\ref{Final_V_eq}, with the output forces and velocities for the actuators ($F_{1}$, $F_{2}$, $V_{1}$ and $V_{2}$) and for the effector along the y and z axes ($F_{y}$, $F_{z}$, $V_{y}$ and $V_{z}$). Equation~\ref{raf} gives the kinematics between joints A, C, and the effector, with the jacobian \textit{$J_{dk}$}. 

\begin{equation}
 \label{raf}
    J_{dk}=\left[\begin{matrix}
    \frac{\partial E_{y}}{\partial q_{A}} & \frac{\partial E_{y}}{\partial q_{C}}\\
    \frac{\partial E_{z}}{\partial q_{A}} & \frac{\partial E_{z}}{\partial q_{C}}
    \end{matrix}\right]
\end{equation}

$E_{y}$ and $E_{z}$ are the y-axis and z-axis positions of the effector, while $q_{A}$ and $q_{C}$ are the angular positions of the rotating joints A and C as illustrated in figure~\ref{fig:model}. Then, equation~\ref{tau_dk} gives the desired forces at the effector.

\begin{equation} 
    \label{tau_dk}
    \left[\begin{matrix}
    \tau_{A}\\
    \tau_{C}
    \end{matrix}\right] = {J_{dk}}^{T}\left[\begin{matrix}
    F_{y}\\
    F_{z}
    \end{matrix}\right]
\end{equation}

where $\tau_{A}$ and $\tau_{C}$ are the torques at the rotating joints A and C. Equation~\ref{act_jac} gives the actuator jacobian \textit{$J_{act}$}, i.e., the kinematics between the actuators and joints A and C.

\begin{equation}
    \label{act_jac}
    J_{act}=\left[\begin{matrix}
    \frac{\partial L_{1}}{\partial q_{A}} & \frac{\partial L_{1}}{\partial q_{C}}\\
    \frac{\partial L_{2}}{\partial q_{A}} & \frac{\partial L_{2}}{\partial q_{C}}
    \end{matrix}\right]
\end{equation}

where $L_{1}$ and $L_{2}$ are described in figure~\ref{fig:model}. The relation between the joint torques and the effector forces is then given by equation~\ref{tau_act}.

\begin{equation}
    \label{tau_act}
    \left[\begin{matrix}
    \tau_{A}\\
    \tau_{C}
    \end{matrix}\right] = {J_{act}}^{T}\left[\begin{matrix}
    F_{1}\\
    F_{2}
    \end{matrix}\right]
\end{equation}
Combining equations~\ref{raf} and \ref{act_jac} yields the total jacobian \textit{$J_{tot}$}.

\begin{equation}
    J_{Tot} = J_{dk}{J_{act}}^{-1}
    \label{Jtot}
\end{equation}

This is the kinematic relationship between the actuators and the effector. Equation~\ref{Final_T_eq} gives the required forces at actuators given the force requirements at the effector.

\begin{equation}
    \label{Final_T_eq}
    \left[\begin{matrix}
    F_{1}\\
    F_{2}
    \end{matrix}\right] = (J_{Tot})^{T}\left[\begin{matrix}
    F_{y}\\
    F_{z}
    \end{matrix}\right]
\end{equation}

For transfer mode, since the first DOF is locked ($q_{A}$ remains constant), $J_{Tot}$ is different. Equation~\ref{Final_V_eq} gives the output velocity at actuator 2 required for the targeted z-axis effector velocity.

\begin{equation}
    \left[\begin{matrix}
    V_{2}
    \end{matrix}\right] = \left[\begin{matrix}
    \frac{\partial L_{2}}{\partial q_{C}}\end{matrix}\right]\left[\begin{matrix}
    \frac{\partial E_{z}}{\partial q_{C}}\end{matrix}\right]^{-1}\left[\begin{matrix}
    V_{z}
    \end{matrix}\right]
    \label{Final_V_eq}
\end{equation}

All those equations will be used in the next sections for designing the actuators and for controlling the prototype.

\subsection{Actuator Design}
\label{Act_des}
A detailed description of the mechanical designs of both actuators 1 and 2 are presented below, followed by a summary of the actuators' output capabilities. 

\subsubsection{Actuator 1}
\begin{figure}[h]
    \centering
    \includegraphics[width=0.9\linewidth]{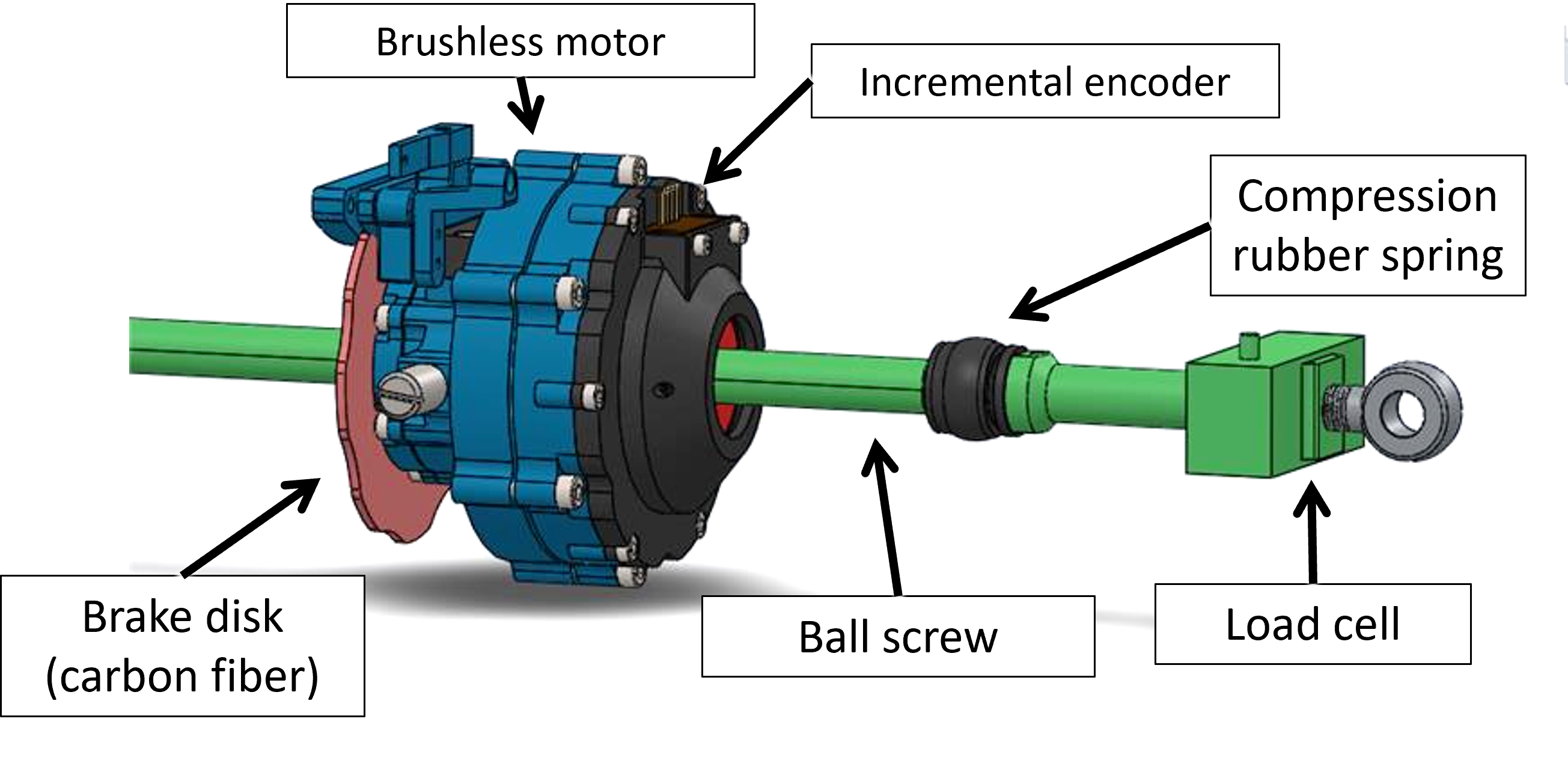}
    \caption{\footnotesize Overall design of actuator 1 with listed components}%
    \label{I_act}%
\end{figure}

is a custom lightly-geared linear actuator as presented in figure~\ref{I_act}.
A high lead 12~mm ball screw (stroke 220~mm, lead 10 mm/turn) is integrated into a frameless BLDC motor (Cubemars Ri70) with a custom aluminum frame. An incremental encoder (US digital EM2 2048~PPR) measures the output angular velocity. For transfer tasks, a carbon fiber disc brake is mounted behind the motor for high breaking force. A S-type load cell (Futek UT4 500 lb) measures the force between the end of the ball screw and joint B. The load cell is not used for the force controller but was used to acquire experimental data for the actuator friction compensation.\\

\subsubsection{Actuator 2 (Dual-Speed Dual-Motor) \cite{LM2S}} 

\begin{figure}
    \centering
    \includegraphics[width=0.9\linewidth]{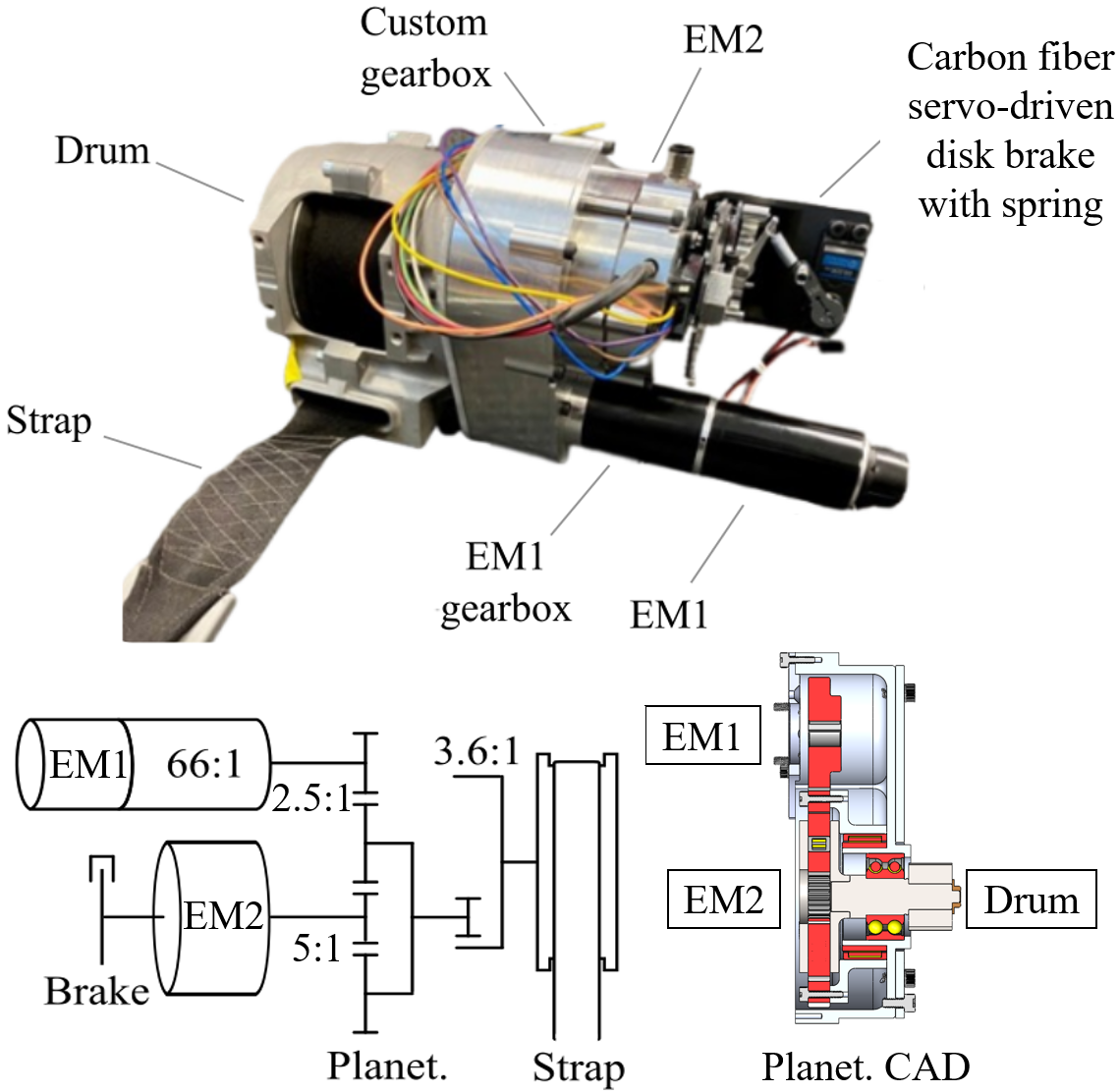}
    \caption{\footnotesize Overall design of actuator 2 with listed components~\cite{LM2S}}%
    \label{LM2S}%
\end{figure}

For this joint, this project uses the same dual-speed dual-motor actuator as developed and tested for a multifunctional ceiling lift~\cite{LM2S} with the architecture shown in figure~\ref{LM2S}. The output belt actuates linkage CDE with a pulley system that doubles the pulling force applied at joint D.

The actuator consists of two motors connected to the belt through a planetary differential gearbox. The first motor is a Maxon RE40 150~W with a 66:1 integrated gearbox for a high force (HF) output used in transfer mode. The second motor is a Tecnotion QTR-A-78-25, which is directly coupled to the sun gear of the gearbox, for a high-speed and low-inertia output used in the force control assisting modes. More design details are available in the paper \cite{LM2S}. Similar to actuator 1, an S-type load cell (Futek UT4 500 lb) measures the belt tension force for experimental purposes. Using this actuator would allow reusing the fall prevention algorithms developed in the ceiling-based project\cite{LM2S}.

\subsubsection{Summary}

The theoretical output specifications of both actuators are presented in table~\ref{table:specs}.
\begin{table}
    \centering
    \caption{Actuators output capabilities at connection points B and D}
    \begin{tabular}{lccc}
        \hline
        & \multirow{2}{*}{Actuator 1} & Actuator 2  & Actuator 2 \\
        & & (high force) & (high speed) \\ \hline
        Ratio (rad:mm) & 0.63:1 & 16.7:1 & 0.45:1   \\ \hline
        \multirow{2}{*}{Max force (N)} & 402 & \multirow{2}{*}{3120} & 579 \\
        & (1725 peak) & & (981 peak)\\ \hline
       Velocity (m/s) & 0.72 & \multirow{2}{*}{0.05} & 0.55   \\ 
        @ load & (0.4 @ peak) & & (0.34 @ peak)\\ \hline
    \end{tabular}
    \label{table:specs}
\end{table} 
In the force-controlled configuration, actuator 1 and actuator 2 (HS configuration) are both used to deploy forces at the effector. In the speed-controlled configuration, actuator 1 is locked using its brake, and only actuator 2 (HF configuration) is used to move the effector.

\begin{figure}
    \centering
    \begin{subfigure}{0.8\linewidth}
        \centering
        \includegraphics[width=1.0\linewidth]{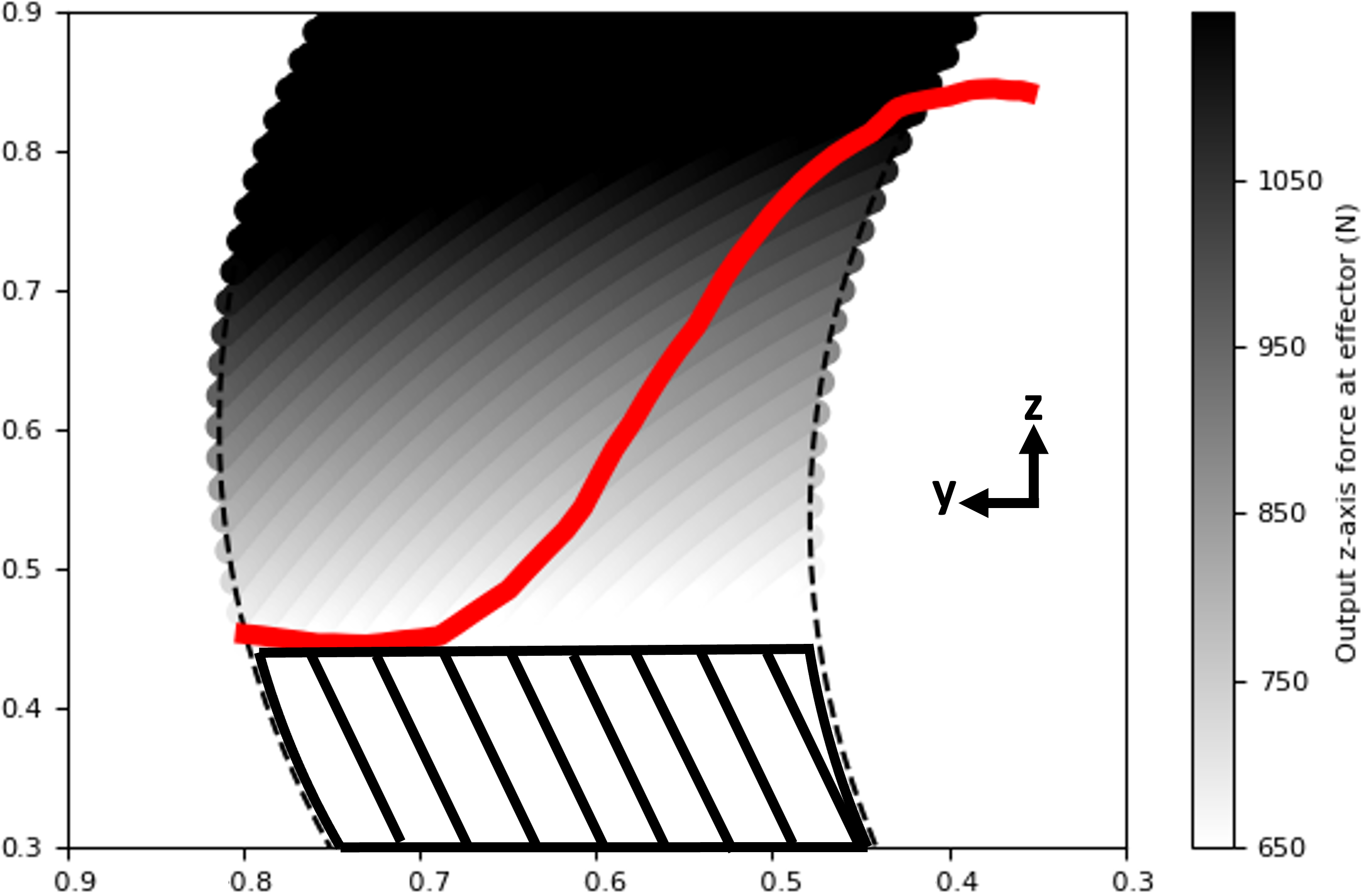}
        \caption{\footnotesize Rehabilitation peak effector force in the workspace (force-controlled configuration). The red line is the typical STS CoM trajectory for an adult male with a stature of 6'3.5".}
        \label{output_rehab}%
    \end{subfigure}
    \\
    \begin{subfigure}{0.8\linewidth}
        \centering
        \includegraphics[width=1.0\linewidth]{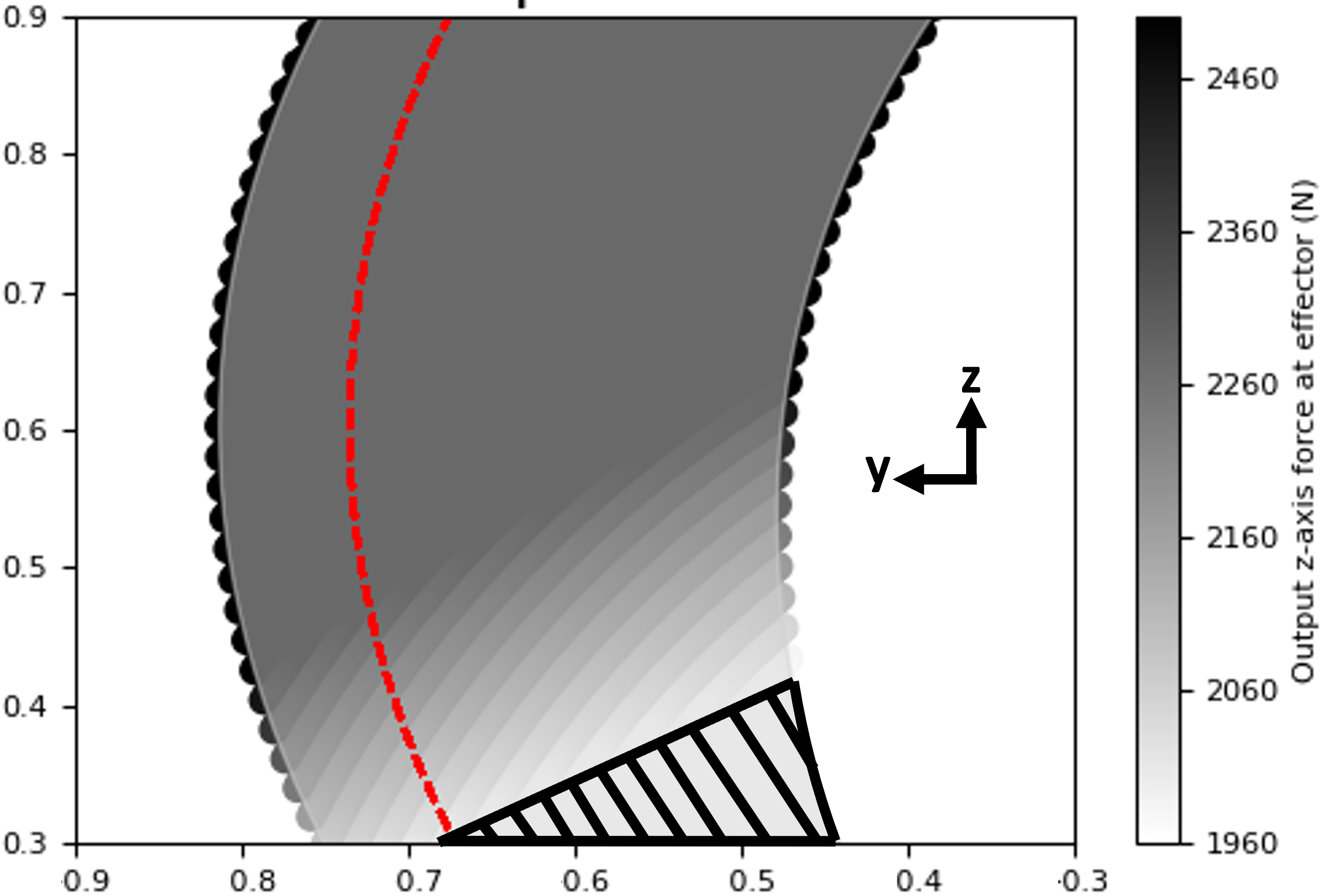}
        \caption{\footnotesize Transfer peak effector force in the workspace (speed-controlled configuration). The red line is an example of a one-DOF transfer trajectory of the CoM.}
        \label{output_transfer}%
    \end{subfigure} \\
    \caption{\footnotesize Effector z-axis peak force capability map with the robot in force control configuration (a) and speed-controlled configuration (b). The black hatched zones represent the area of the workspace in which the effector is out of the output z-axis force specification for rehabilitation and transfer assisting modes (section~\ref{Sys_req}).}
\end{figure}

Figure~\ref{output_rehab} presents the peak output z-axis force at the effector in the whole workspace for the force-controlled configuration.
For example, the red curve represents the effector trajectory for an STS motion of a healthy adult male with a stature of 1.91~m using the custom harness~\cite{Joint_angles,Percentile}. This is approximately the largest y and z axes CoM displacements expected in a clinical setting with the data available. 
As displayed in figure~\ref{output_rehab}, the output z-axis force at the effector is over the minimum requirement of 650~N for rehabilitation from section~\ref{Sys_req} in the workspace, except for a zone below 0.45~m in the z-axis. Regarding the displayed trajectory, the end point is slightly outside of the workspace highlighting that a slightly longer ball screw should have been used for actuator 1. This limitation will be investigated in section~\ref{experiments}. 
The available effector velocities under load are higher than the requirements of table~\ref{Sum_req}. Similarly, both actuators at peak output force allow the effector to meet the y-axis output force requirement in the workspace. 
Figure~\ref{output_transfer} presents the peak output z-axis force at the effector for the whole workspace in the speed-controlled configuration using peak output force.


As displayed in figure~\ref{output_transfer}, the output z-axis force at the effector is over the minimum requirement of 1962~N for transfer mode from section~\ref{Sys_req} in the workspace, except for a zone in the bottom right corner. This zone can be avoided by braking (locking) actuator 1 near its maximal length. The velocity of actuator 2 at the effector load described in table~\ref{Sum_req} for the speed-controlled configuration is sufficient to meet the transfer requirement described in section~\ref{Sys_req}. Section~\ref{experiments} presents experimental results with weights on the effector to validate these results.

\subsection{Controller Design} \label{Algo}
\subsubsection{Rehabilitation Modes Force Controllers (Mobility Levels A-C)}
\begin{figure}
    \centering
    \includegraphics[width=\linewidth]{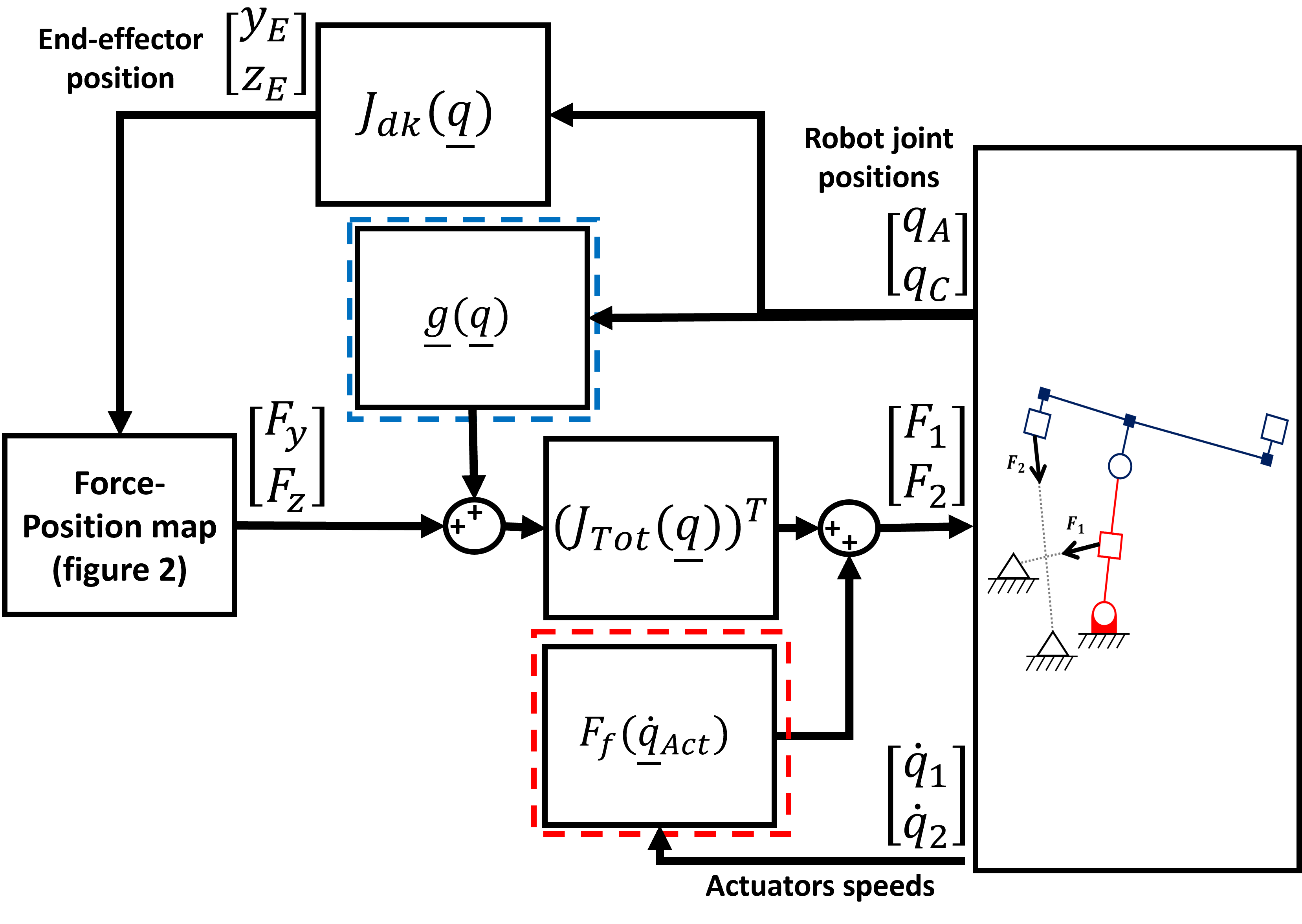}
    \caption{\footnotesize Layout of the force controller. The blue dashed zone is the structure mass compensation while the red dashed zone is the actuator friction compensation.}
    \label{Ctrl_F}
\end{figure}

As described in the section~\ref{Sys_req}, the assisting modes from Follow Me to CoM Balance use a controller which converts the desired output forces at the effector ($F_{yd}$ and $F_{zd}$) into command output forces for the actuators ($F_{1}$ and $F_{2}$).
An open-loop force controller was chosen for this project for its robustness and simplicity since it does not rely on any force measurement. 
The force measurements are used for validation purposes only.
The desired y and z-axis forces are determined by the force-position map of the assisting mode in use, presented in figure~\ref{Ctrl_modes}.
These forces are fed to the controller as an input ($F_{yd}$ and $F_{zd}$). Then, $F_{zd}$ is augmented to compensate for the robot structure's mass. Once the desired actuator output forces ($F_{1d}$ and $F_{2d}$) are calculated using equation~\ref{Final_T_eq}, the output forces are augmented to compensate for each actuator's internal friction. 
These friction compensations are calculated according to the friction model of each actuator and their respective rotational speed measured by the encoders on the BLDC motors. The final output forces for each actuator ($F_{1}$ and $F_{2}$) are then sent to each actuator's BLDC controller. The detailed layout of this controller is presented in figure~\ref{Ctrl_F}.

Variables $q_{A}$ and $q_{C}$ are the angular positions of joints A and C. Variables $\dot{q_{1}}$ and $\dot{q_{2}}$ are the rotational velocities of the BLDC motor in actuator 1 and actuator 2 respectively, with $\dot{q_{Act}}$ representing the column vector of the rotational velocities of both actuators. $y_{E}$ and $z_{E}$ are the cartesian coordinates of the harness anchor point using the direct kinematics Jacobian ($J_{dk}$). The blue zone is the structure mass compensation while the red zone is the actuator friction compensation.

\begin{figure}
    \centering
    \begin{subfigure}[b]{0.48\linewidth}
        \centering
        \includegraphics[width=\linewidth]{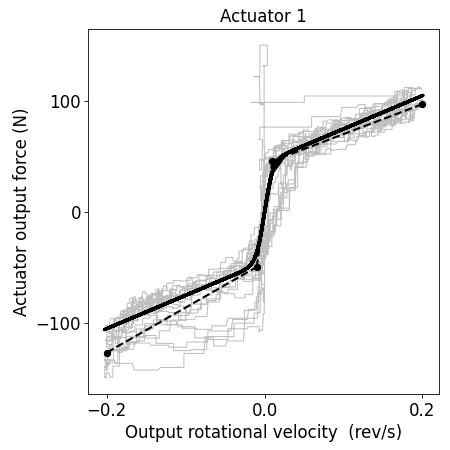}
        \caption{\footnotesize Actuator 1}
    \end{subfigure}
    \begin{subfigure}[b]{0.48\linewidth}
        \centering
        \includegraphics[width=\linewidth]{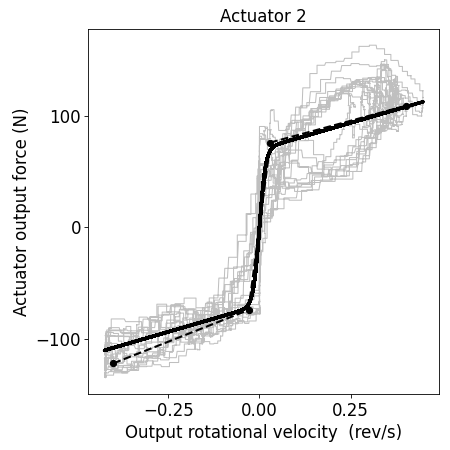}
        \caption{\footnotesize Actuator 2}
    \end{subfigure} \\
    \caption{\footnotesize Friction model of both actuators with linear approximation (dashed line) and $tanh$ approximation (solid line)}
    \label{fric_model}
\end{figure}

The \textbf{actuator friction compensation} accounts for the dry and viscous friction for a more precise effector force. 
Data for the models are acquired by imposing a zero torque command and then moving the output in a cyclic motion at varying speeds. Output forces read at the load cells are then plotted according to the electric motor rotational velocities, as shown in figure~\ref{fric_model}.
The dry and viscous friction data are then approximated with a hyperbolic tangent function $F_{f} = a\cdot \textrm{tanh}(b\cdot \dot{q}_{Act})$ (where $Act$ is either $1$ or $2$ depending on the actuator), shown by the black lines in figure~\ref{fric_model}. 

The \textbf{linkage mass compensation} supports the static frame mass to increase the precision of the effector force. The required forces at the effector to compensate for the mass model are calculated with equations~\ref{eq_g_prelim} and~\ref{eq_g}.

\begin{equation}
\begin{split}
V(\underline{q}) &= m_{h}g(L_{h}cos(q_A)) + \\
 & m_{v}g(L_{AC}cos(q_A) + L_{CD}cos(q_A) - L_{v}sin(q_A+q_C))
\label{eq_g_prelim}
\end{split}
\end{equation}

\begin{equation}
    \underline{g}(\underline{q}) = \frac{\partial{V(\underline{q})}}
    {\partial{{\underline{q}}^T}} = 
    \begin{bmatrix}
    \frac{\partial{V}}{\partial{q_A}}\\\\
    \frac{\partial{V}}{\partial{q_C}}
    \end{bmatrix}
    \label{eq_g}
\end{equation}

In equation \ref{eq_g_prelim}, $m_h$ and $m_v$ are the masses of the horizontal and vertical linkages, while $L_h$ and $L_v$ are the distances between the linkage masses and joints A and C respectively. The conservative force vector is then integrated in the robot controller to add to the command effector forces.

\subsubsection{Transfer Mode Speed Controller (Mobility Level D)} 
\begin{figure}
    \centering
    \includegraphics[width=0.9\linewidth]{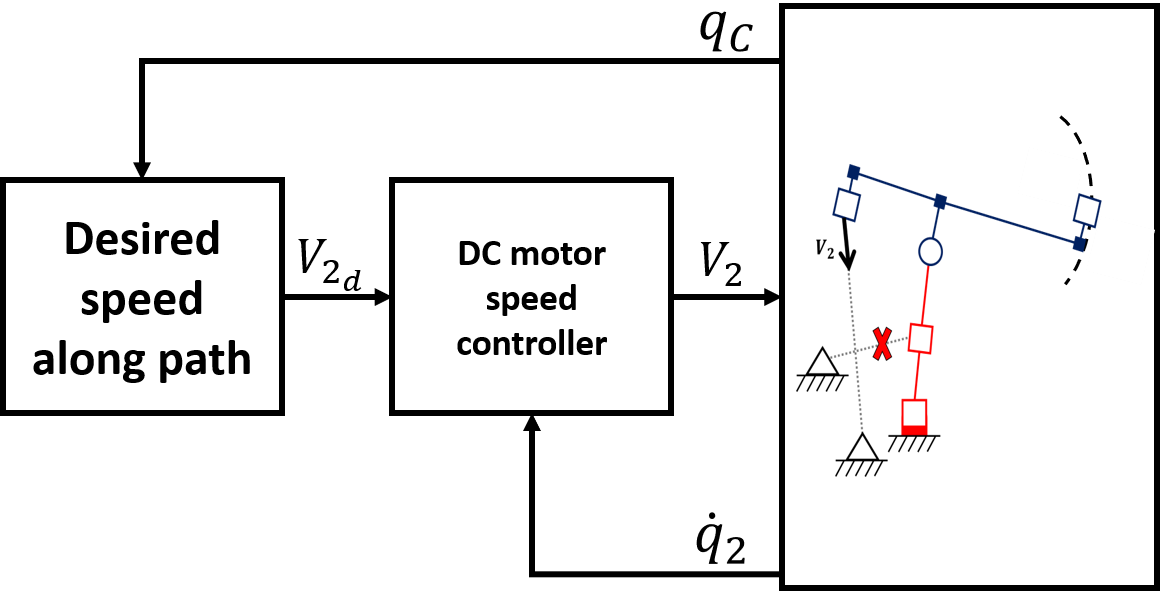}
    \caption{\footnotesize Layout of the speed controller}
    \label{Ctrl_V}
\end{figure}

Figure~\ref{Ctrl_V} presents the effector speed controller layout used for the transfer tasks.
$V_{z}$ is set to 0.3~m/s to reproduce commercial raising aid devices, as specified in section~\ref{Sys_req}. Then, equation \ref{Final_V_eq} gives the desired output velocity of actuator~2 in HF configuration ($V_{2_{d}}$) as a function of the angular position of joint C ($q_{C}$). The actuator reaches this speed reference with a velocity feedback controller using the incremental encoder included on the Maxon RE40 motor.

\section{EXPERIMENTS}
\label{experiments}
\begin{figure}
    \centering
    \includegraphics[width= 0.75\linewidth]{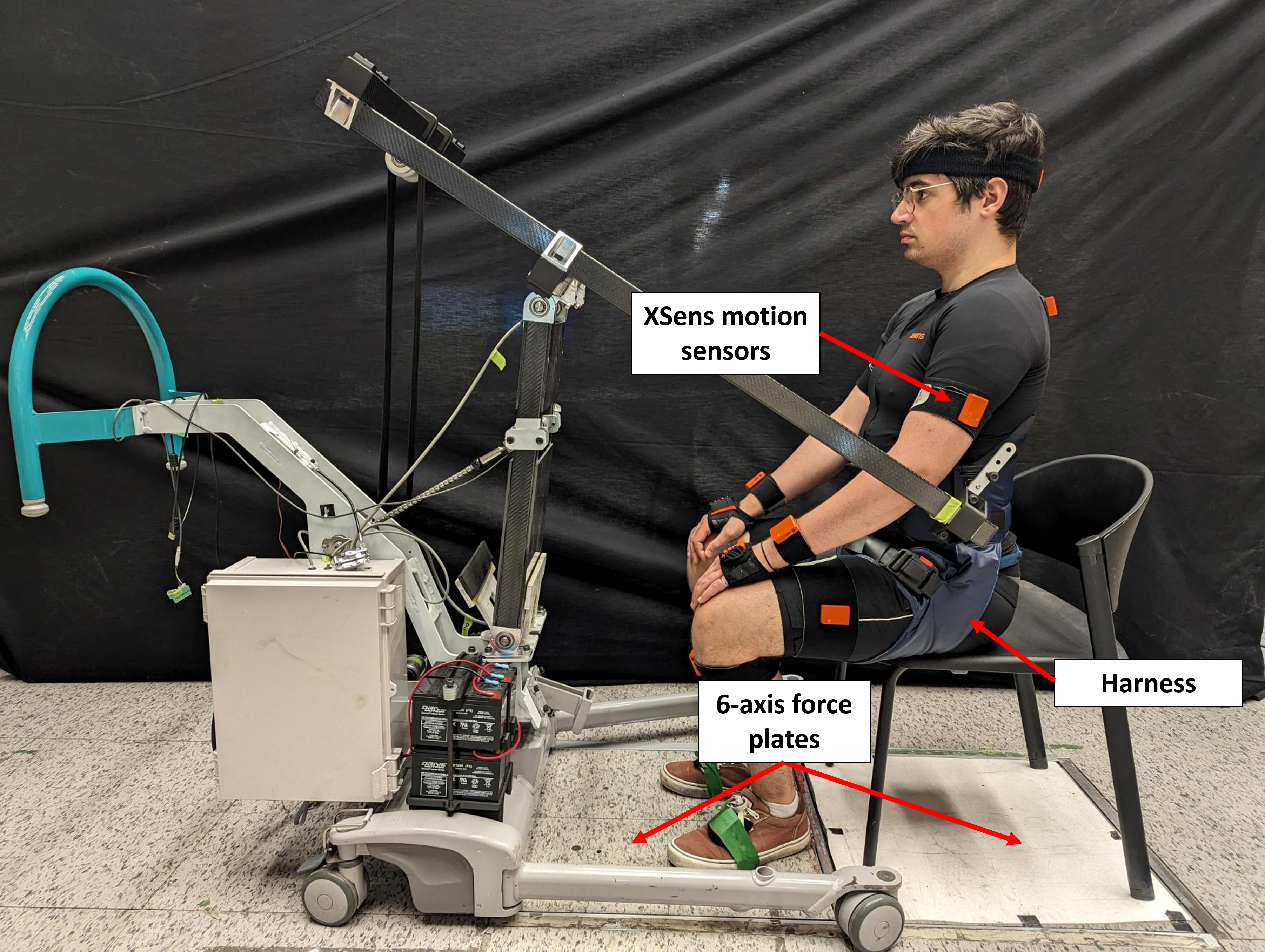}
    \caption{\footnotesize Experimental setup}
    \label{test_setup}
\end{figure}

For all the experiments, ground reaction forces (GRFs) and CoM velocity/acceleration data are acquired to analyze the impact of the different assisting modes on key characteristics of the STS motion.
GRFs are measured with two six-axis AMTI OR6 Series force plates recessed in the floor, one for the participant's feet and the other for the chair. 
GRFs data is acquired at 2000~Hz, using AMTI load cell conditioners and a Vicon MX Ultranet HD acquisition module.
CoM velocity and acceleration data are acquired at 60~Hz using a XSens Awinda inertial motion tracking system.
The data from the motion tracking system is then used to create a model of the participant's body with its segments, based on the International Society of Biomechanics.
The position of the user's CoM is estimated using a point calculated by the software near the L3 vertebrae. Further information on the model used can be found in section 23.2 of the MVN User Manual by Xsens~\cite{MVN_User_Manual}.
To avoid arm strength to have an impact on STS motion, an armless plastic chair with a seat height of 0.43~m is used.
Users were instructed to sit on the chair and to connect themselves to the robotic device via a harness. 
Figure~\ref{test_setup} presents the typical setup for the STS motion tests.

Eight healthy adults (50\% male and 50\% female) participated in this study (age: 27.13$\pm$4.07 years, height: 1.75$\pm$0.04 m, weight: 81.13$\pm$12.86 kg). Participants can perform multiple STS with no assistance. Due to the limited stroke of actuator~1, participants were unable to rest their back on the chair's backrest at the start of each repetition. The experimental protocol was approved by the local ethics committee (University of Sherbrooke, CÉR-LSH 2024-4304, approbation date: 2022-12-12) and the participants all provided informed consent.

\subsection{Robot Arm Transparency in Follow Me Mode}
\begin{figure}
    \centering
    \includegraphics[width= 0.9\linewidth]{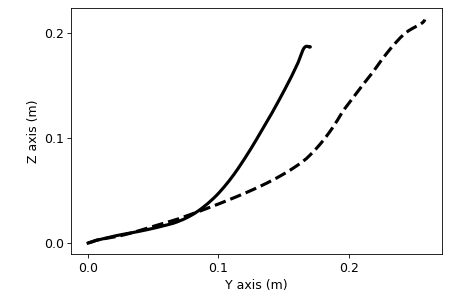}
    \caption{\footnotesize WR (solid) and WoR (dashed) CoM trajectories during STS motion (All participants, normalized to user height)}
    \label{Ares_pos}
\end{figure}

The goal of this experiment is to compare peak value of key STS metrics when users use the robot in Follow Me mode to their unhindered STS motion. To do so, all eight participants were asked to perform two sets of five repetitions of STS motion, one with the harness but without robot (WoR) and one with the harness and with robot (WR) set in Follow Me mode. The participants were asked to perform the repetitions at a self-selected speed, with a pause of two seconds between each repetition to make sure participants come to a stop. CoM displacement results indicate that participants have, on average, lower y-axis and z-axis displacements in the WR scenario compared to the WoR scenario, as shown in figure~\ref{Ares_pos}. 

\begin{figure}
    \centering
    \includegraphics[width= \linewidth]{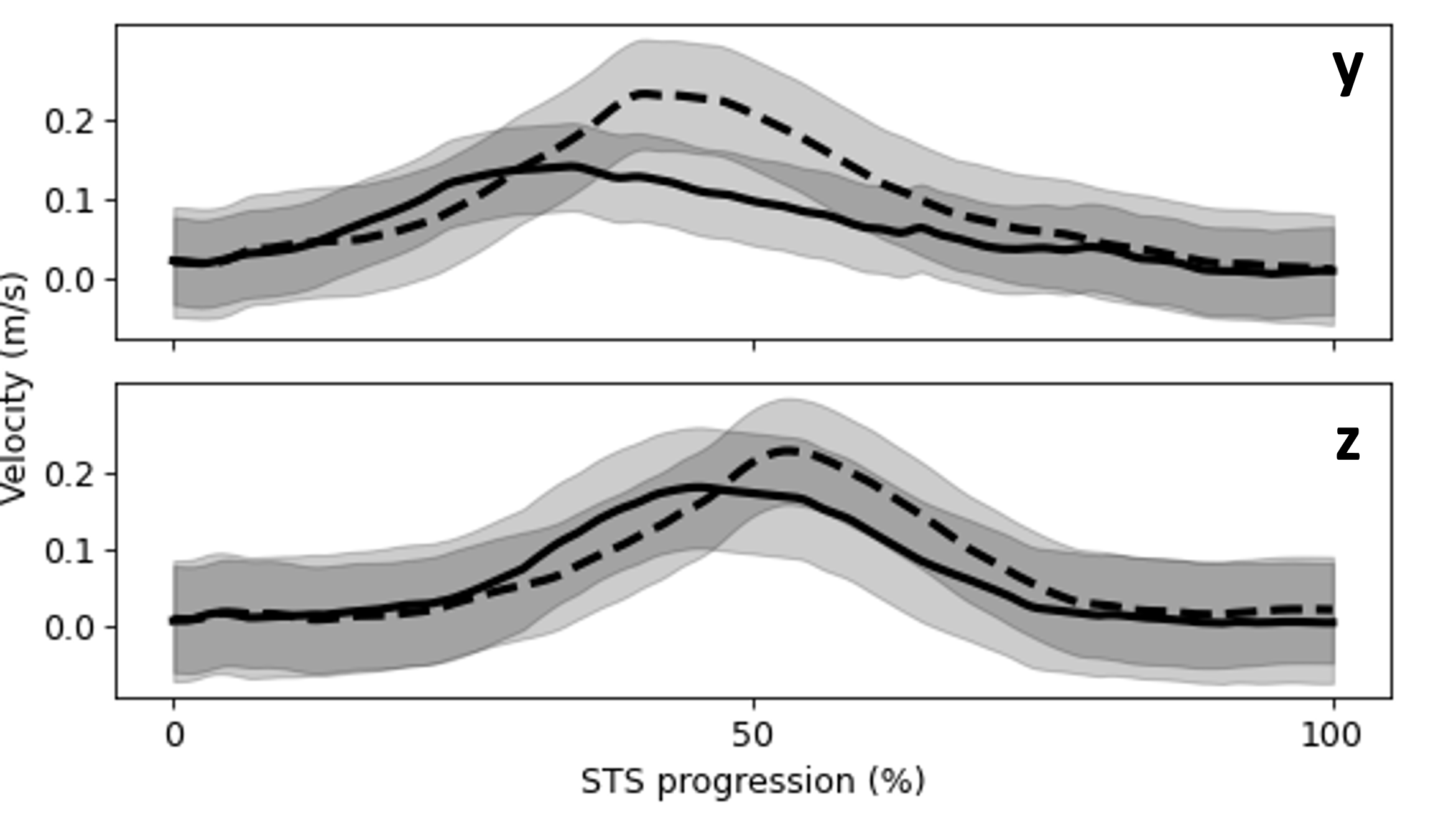}
    \caption{\footnotesize WR (solid) and WoR (dashed) CoM velocities during STS motion (All participants, normalized to user height)}
    \label{A_res_vel}
\end{figure}

Each participant's repetitions used to build the WR and WoR averages are normalized with their stature to give a uniform reference. Average peak values of CoM velocity/acceleration and GRFs (normalized to each participant's bodyweight) in both the y and z axes are then calculated for WR and WoR sets. Using a signed bilateral Wilcoxon statistical test ($\alpha$ = 0.05), the average peak value of both sets is then compared for each participant to detect any bias. The results indicate that only the y-axis and z-axis CoM peak velocity are consistently impacted by mode A, with a reduction of 28.1\% and 25.9\% respectively, compared to WoR. Figure~\ref{A_res_vel} shows an example of the impact of Follow Me mode on the CoM velocities. 

\begin{figure}
    \centering
    \includegraphics[width= 0.9\linewidth]{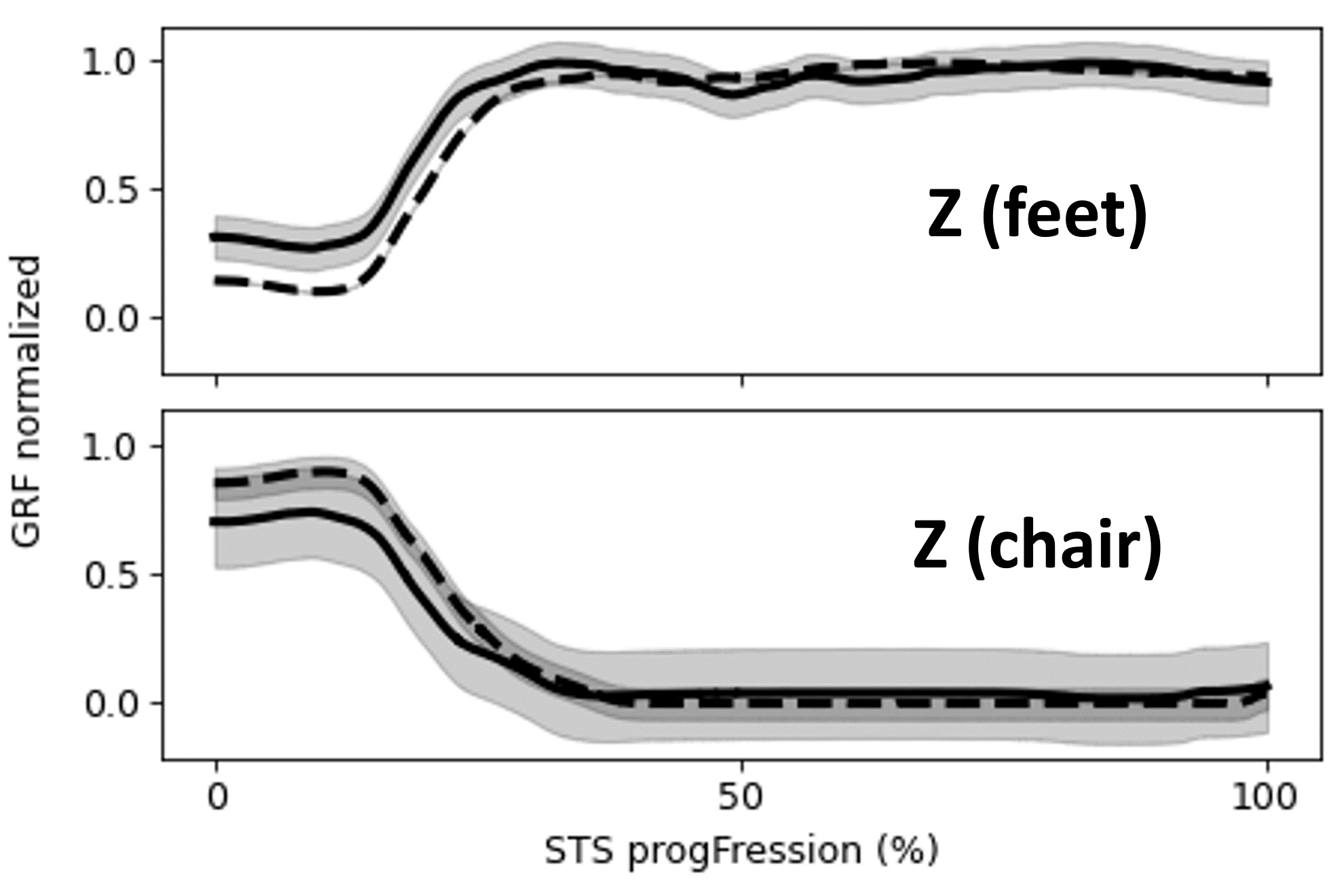}
    \caption{\footnotesize WR (solid) and WoR (dashed) z-axis GRFs during STS motion (All participants, normalized to user weight)}
    \label{A_res_F}
\end{figure}

CMC analysis \cite{CMC} to compare WR and WoR velocity curves in both the y and z axes gives 84\% and 81\% similarity (on average) respectively. No significant difference between WR and WoR is detected for peak CoM acceleration and GRFs (p $<$ 0.05). However, the GRF data indicates that the distribution of the participants' bodyweight between the chair and the feet varies between the WR and WoR datasets. 
The GRFs results for this experiment are presented in figure~\ref{A_res_F}.

There was no notable difference in the y-axis GRF between all sets of five STS.

\subsection{Force Fidelity for Weight Unloading Mode}
As defined in section~\ref{Sys_req}, the Weight Unloading mode requires that the z-axis output force at the effector is within a 5\% error to the input desired force to the force controller. Furthermore, since STS is a complex motion, the controller must be able to respect the requirement at various speeds. To verify that the prototype meets this requirement, experiments were done to compare the vertical assistance applied on the user's body during STS motions to the desired vertical assistance. All eight participants performed four sets of five repetitions of STS motion, with each set at 0\%, 5\%, 10\% and 20\% target vertical assistance respectively. The participants were asked to perform the repetitions at self-selected speed, with a pause of two seconds between each repetition to make sure participants come to a stop. The order of each set was randomized, and the assistance level was kept unknown to participants. 
For each repetition, the applied vertical force on a participant is measured by adding both the vertical (z-axis) GRFs for the chair and the feet force plates. By combining the data of all participants for each set, a dataset of 40 repetitions is created for every target assistance level.
Each repetition's force data is therefore normalized to its respective participant's bodyweight, into dimensionless measured assistance. Each normalized dataset is then averaged to obtain the mean measured assistance $As_{mean}$, using equation~\ref{W_norm}, which is compared to its target assistance level.

\begin{equation}
    As_{mean} = \frac{1}{8}\sum_{j=1}^{8}\frac{\sum_{i=1}^{5}Fz_{ ij}}{Weight_{ j}}
    \label{W_norm}
\end{equation} 

$Fz_{ ij}$ is the sum of the vertical measured forces for the chair and feet force plates at STS repetition $i$ for participant $j$. $Weight_{ j}$ is the bodyweight of participant $j$. Table~\ref{B_test_table} presents these results and the non-normalized force error for each assistance level.

\begin{table}
    \centering
    \caption{Force fidelity experiment results (all units are in \%~bodyweight of users)}
    \begin{tabular}{cc}
    \hline
        Target Assistance & Measured Assistance Error
        \\\hline
         $0\%$ & $0.17\%\pm2.43$ \\\hline
         $5\%$ & $1.27\%\pm1.95$ \\\hline
         $10\%$ & $0.5\%\pm3.56$ \\\hline
         $20\%$ & $-0.19\%\pm5.70$\\\hline
    \end{tabular}
    \label{B_test_table}
\end{table}

For each trial except the 5\% target assistance, the mean error between the measured assistance level and the desired assistance level is within 5\% (p $<$ 0.05). However, the standard deviation in the data for each target assistance level exceeds the 5\% force fidelity requirement, which means that there is significant variability in the results. When performing one-sided statistical t-tests ($\alpha$=0.05) for each target assistance level, the null hypothesis that the assistance error does not exceed 5\% for each assistance level is not rejected. Furthermore, no correlation between user anthropometric data (height, weight, age and sex) and average assistance error was observed in the experimental data.

\subsection{Impact of Virtual Spring in CoM Balance Mode}
\begin{figure}
    \centering
    \includegraphics[width=0.9\linewidth]{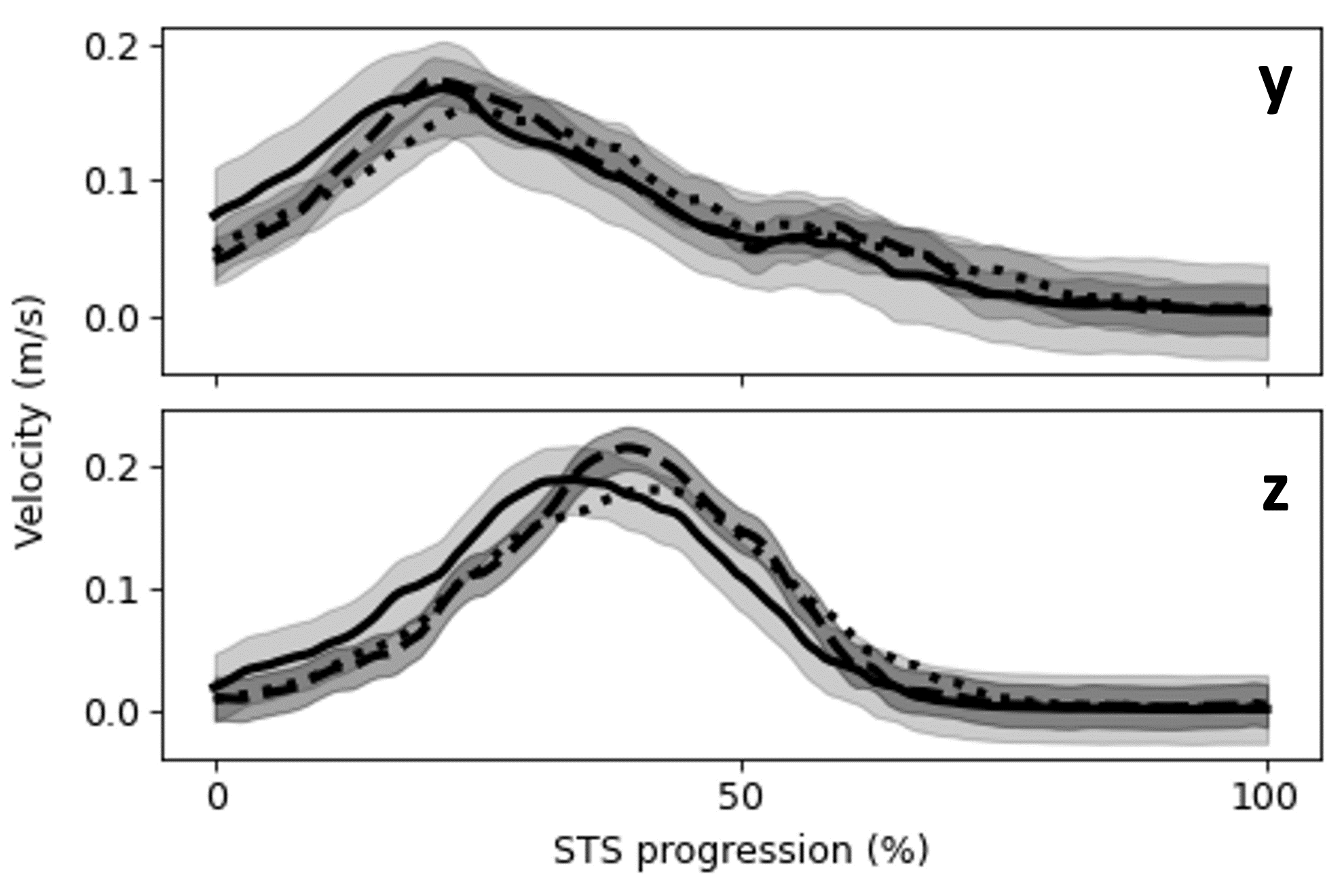}
    \caption{\footnotesize Average CoM velocities normalized to users' heights: 0~N/m (solid), 200~N/m (dashed), 300~N/m (dotted)}
    \label{C_test_vel}
\end{figure}

For the CoM Balance mode validation, the goal is to evaluate the effect of adding a y-axis virtual spring to the z-axis constant force of the Weight Unloading mode. Eight participants completed three sets of five STS repetitions, each with a different y-axis virtual spring rigidity (0~N/m, 200~N/m and 300~N/m). The participants performed the repetitions at self-selected speed, with a two-second pause each time to ensure a full stop.
As in the Weight Unloading mode, the set order was randomized and unknown to the participants. Additionally, a 5\% bodyweight unloading assistance was used to compare to the Weight Unloading mode and keep an upward tension in the harness. The CoM trajectory, velocity, acceleration, and the GRFs were then compared across the five STS sets. No significant differences were observed in CoM trajectories and accelerations between all three sets. For velocity, the 300~N/m virtual spring led on an average 14\% increase in peak z-axis velocity, without a notable change in the y-axis. The 200~N/m spring showed no significant effect on velocity, as shown in figure~\ref{C_test_vel}.

\begin{figure}
    \centering
    \includegraphics[width=0.9\linewidth]{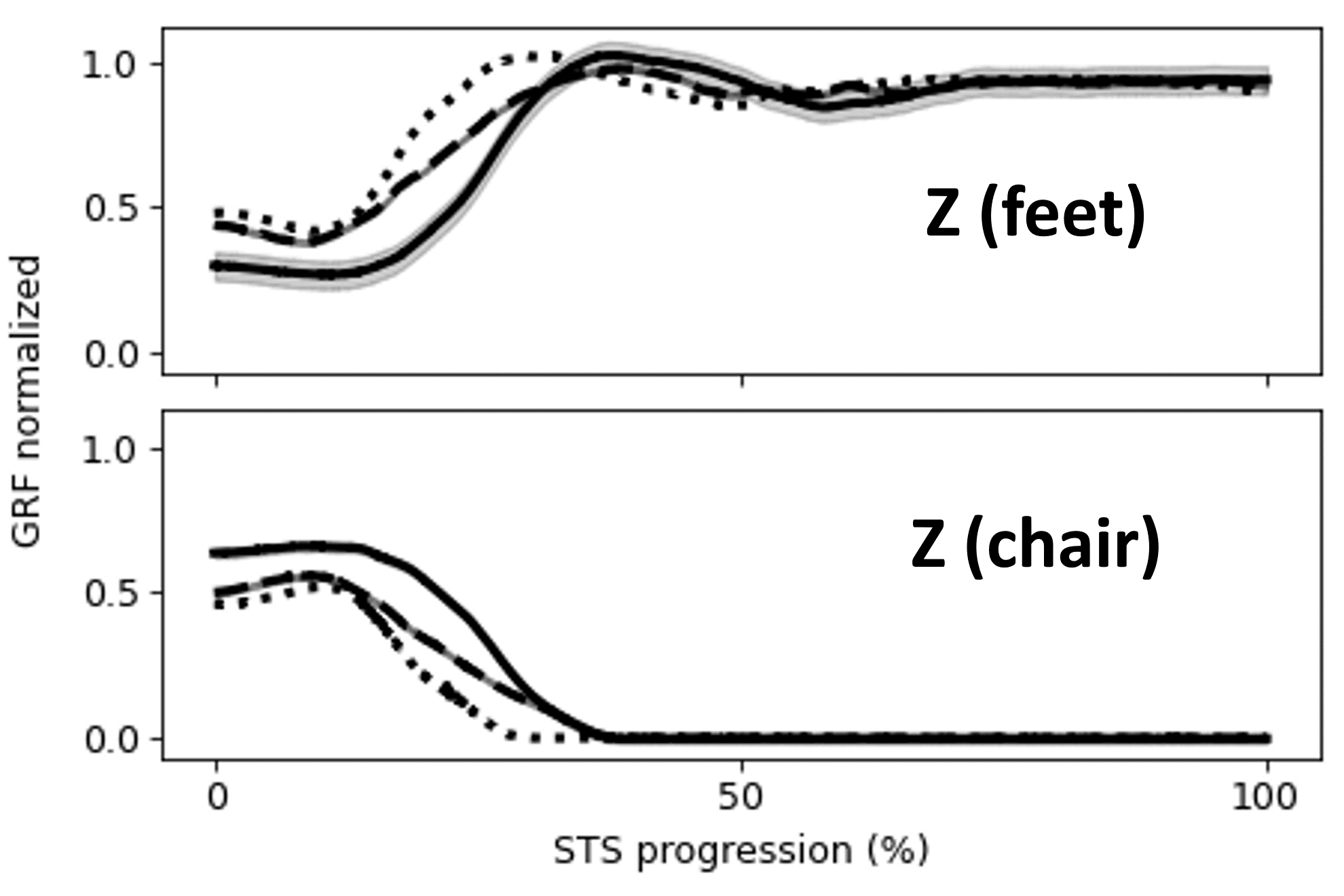}
    \caption{\footnotesize Average z-axis GRFs normalized to users' weights: 0~N/m (solid), 200~N/m (dashed), 300~N/m (dotted).}
    \label{C_test_F}
\end{figure}

Additionally, the GRF data indicates that the distribution of the participants' bodyweight between the chair and the feet varies between each virtual spring rigidity dataset. At the beginning of the motion, before seat-off, participants' feet support on average 14\% and 18\% more of their total bodyweight with the 200~N/m and 300~N/m virtual springs respectively. The average variation of z-axis GRFs during the STS motion is presented in figure~\ref{C_test_F}. 
There is no notable difference in the y-axis GRF between all sets of five STS.

\subsection{Transfer Mode (Mobility Level D)}
The Transfer mode aims to reproduce the assistance from commercially available transfer rising aids by lifting the patient's entire bodyweight on a fixed trajectory. The output vertical force of actuator 2's HF configuration is sufficient to match the transfer specifications, except for a small zone of the workspace. This limitation could, however, be fixed by lowering the entire structure of the robot arm closer to the ground, to make the pulling force from actuator 2 generate a higher joint C torque. Another limitation of the prototype in this mode is the frame strength. Indeed, tests revealed that tubes with thicker walls should be used for the CDE linkage to lift 205~kg. Transfer tests involved lifting and lowering weight plates up to 96~kg. To ensure participants' safety, no transfers were performed on them. Table~\ref{table:transfer} summarizes the velocities reached for each load, which are higher than the 0.03~m/s requirement.

\begin{table}
    \centering
    \caption{Mean speed measured during raising and lowering of weight plates in transfer mode}
    \begin{tabular}{ccc}
    \hline
        Weight (kg) & Lifting speed (m/s) & Lowering speed (m/s)
        \\\hline
         28 & 0.039 & 0.039\\\hline
         51 & 0.040 & 0.040\\\hline
         74 & 0.041 & 0.040\\\hline
         98 & 0.044 & 0.044\\\hline
    \end{tabular}
    \label{table:transfer}
\end{table}

\section{DISCUSSION \& CONCLUSION}
\label{Disc&Concl}
This paper introduces a prototype floor-based robot capable of STS transfer and rehabilitation assistance with various training profiles using a low-inertia two DOFs structure and a dual-motor actuator combined with a linear actuator which can be blocked. The prototype lifts patients up to 200~kg at a vertical (z-axis) speed of over 0.03~m/s in transfer mode. For force control modes, the prototype unloads up to 61~kg at 0.35~m/s in the z-axis while providing up to 18~kg of assistance at 0.33~m/s in the y-axis. 

\subsection{Follow Me Mode (Mobility Level A)}
The Follow Me mode aims to allow a patient to perform an STS on their own, with no assistance and no restraints. Thus, the objective for the Follow Me mode was to demonstrate that the prototype does not impact the natural STS motion. Peak values of key STS metrics in all participants were compared between using no robot and being connected to the robot. The results showed that peak velocity is consistently lower WR than WoR. However, CMC analysis indicates that using mode A has a low impact on the shape of the CoM speed profiles. While a reduction in peak velocity is not desirable for rehab training, it is important to note that no significant difference in peak CoM acceleration or GRFs was detected between both test conditions. This means that participants did not have to deploy more forces to the ground to perform the STS motion but that they did not accelerate for as long WR compared to WoR. A hypothesis for this observation could be that, when participants are constrained in the prototype, they are inherently less comfortable in moving quickly compared to their unhindered motion. Another explanation is that the modified seating position due to the limited stroke of actuator 1 (see sections~\ref{Des&Mod} and \ref{experiments}) means users could not sit fully at the back of the chair and therefore had more weight on their feet (see figure~\ref{test_setup}). This also explains why more of the user's bodyweight is supported by the feet than the chair when using the prototype, since the sitting position is altered. Changing initial foot placement also reduces the forward momentum needed before seat-off\cite{Feet_placement}, which could explain the lower velocity peaks. As stated in section~\ref{Des&Mod}, this could be solved by choosing a different design for actuator 1 or by modifying its position on the prototype to maximize the displacement of the first DOF. Participants reported that they did not feel any noticeable hindrance between WR and WoR trials, which would be consistent with the GRFs results.

\subsection{Weight Unloading Mode (Mobility Level B)}
The Weight Unloading mode aims to allow a patient with insufficient lower body muscle power to perform a STS by helping lift their CoM upward with a constant z-axis effector force. The mean precision of provided assistance is measured to be within the desired 5\% maximum error (except for 5\% target assistance) presented in section~\ref{Sys_req}, however, there is significant variability in the results. While the statistical t-tests do not conclude that the error exceeds 5\%, this might be due to an insufficient amount of data. Whether the variability in the measured assistance is due to participant behavior or lack of precision from the open-loop force controller will need to be investigated further. However, participants did feel that increasing the target assistance in this mode reduced the effort necessary to accomplish the STS motion.

\subsection{CoM Balance Mode (Mobility Level C)}
The CoM Balance mode aims to assist patients who are unable to generate enough forward and upward momentum to perform the STS by providing a virtual spring in the y-axis and a constant force in the z-axis. Surprisingly, the addition of a y-axis virtual spring does not significantly increase CoM y-axis velocity and acceleration. However, this can be accounted more to the participant's behavior toward a forward force than an objective effect of the assisting mode. Otherwise, the results from the experiments indicate that higher virtual spring rigidities lead to a small increase of peak z-axis velocity and cause more of the participants' bodyweight to be supported by their feet before seat-off. For the z-axis peak velocity increases, this can seem counter-intuitive as it is not in the same axis as the virtual spring. An explanation for this observation is due to the increased proportion of the bodyweight supported by the feet before seat-off. Indeed, the trunk swing before seat-off is used to transfer bodyweight over the feet before lifting the CoM by generating forward momentum~\cite{Momentum,Failed_STS}. Since the virtual spring helped this transfer, the participants did not need to generate as much forward momentum using their trunk compared to the "normal" motion, resulting in no significant increase in y-axis peak velocity. However, during the experiment, all participants reported that the addition of a y-axis virtual spring made the STS motion easier compared to only a z-axis constant force, such as the Weight Unloading mode, especially at higher rigidity levels.

\subsection{Future Work}
The goal of this research is to demonstrate the potential of a two DOFs multifunctional STS assistance robot to adapt training to various patient needs. Experiments show that, by changing the force controller input values as well as a reconfigurable architecture, it is possible to assist both seat-off and CoM lifting while maintaining transfer capabilities for low-mobility patients. Having such a system in hospitals or clinical care centers would allow staff to assist various patient profiles, without resorting to different types of lifts and having to move patients to dedicated training rooms. Future interests for the prototype would be to do clinical trials to test the assisting modes with real patients of various mobility levels, in order to refine the force field maps from figure \ref{Ctrl_modes}.



\bibliographystyle{elsarticle-num}
\bibliography{ref}

@article{Changesinmusclepower, title={Changes in muscle power after usual care or early structured exercise intervention in acutely hospitalized older adults}, volume={11}, ISSN={2190-5991, 2190-6009}, DOI={10.1002/jcsm.12564}, year={2020}, month={Aug}, pages={997–1006}}

@article{Motorandcognitive, title={Motor and Cognitive Functional Status Are Associated with 30-day Unplanned Rehospitalization Following Post-Acute Care in Medicare Fee-for-Service Beneficiaries}, volume={31}, ISSN={0884-8734, 1525-1497}, DOI={10.1007/s11606-016-3704-4}, number={12}, journal={Journal of General Internal Medicine}, author={Middleton, Addie and Graham, James E. and Lin, Yu-Li and Goodwin, James S. and Bettger, Janet Prvu and Deutsch, Anne and Ottenbacher, Kenneth J.}, year={2016}, month={Dec}, pages={1427–1434}}

@article{Functionalstatus, title={Functional status before and during acute hospitalization and readmission risk identification}, volume={11}, ISSN={1553-5592, 1553-5606}, DOI={10.1002/jhm.2595}, number={9}, journal={Journal of Hospital Medicine}, author={Tonkikh, Orly and Shadmi, Efrat and Flaks‐Manov, Natalie and Hoshen, Moshe and Balicer, Ran D. and Zisberg, Anna}, year={2016}, month={Sep}, pages={636–641}}

@article{Barriers_EM, 
        author = {Dubb, Rolf and Nydahl, Peter and Hermes, Carsten and Schwabbauer, Norbert and Toonstra, Amy and Parker, Ann M. and Kaltwasser, Arnold and Needham, Dale M.}, 
        title = {Barriers and Strategies for Early Mobilization of Patients in Intensive Care Units}, 
        journal = {Annals of the American Thoracic Society}, 
        volume = {13}, 
        pages = {724–730}, 
        year = {2016}, 
        ISSN = {2329-6933, 2325-6621}, 
        DOI={10.1513/AnnalsATS.201509-586CME}, number={5}
}

@article{ICUAW, title={Clinical review: intensive care unit acquired weakness}, volume={19}, ISSN={1364-8535}, DOI={10.1186/s13054-015-0993-7}, number={1}, journal={Critical Care}, author={Hermans, Greet and Van Den Berghe, Greet}, year={2015}, month={Dec}, pages={274}}

@article{STS_Event_std, 
        author={Etnyre, B. and Thomas, D. Q},
        title={Event Standardization of Sit-to-Stand Movements},
        journal={Physical Therapy},
        volume={87}, 
        number={12},
        pages={1651–1666}, 
        year={2007},
        ISSN={0031-9023, 1538-6724}, 
        DOI={10.2522/ptj.20060378},
}

@article{STS_Traj_model, 
         author = {Shi, Yaochen and Meng, Lingyang and Chen, Daimin},
         title = {Human Standing-Up Trajectory Model and Experimental Study on Center-Of -Mass Velocity}, 
         journal = {IOP Conference Series: Materials Science and Engineering},
         volume = {612}, 
         number = {2},
         pages = {22-88},
         year = {2019},
         ISSN = {1757-8981, 1757-899X}   
}

@article{BMAT2, title={The Bedside Mobility Assessment Tool 2.0}, author={Boynton, Teresa and Kumpar, Dee and VanGilder, Catherine}, language={en},  year={2020},  journal={American Nurse Official Journal} }

@book{PDRAMA,
        author={Kamnik, R. and Bajd, T. and Williamson, J. and Murray-Smith, R.},
        title={Rehabilitation Robot Cell for Multimodal Standing-Up Motion Augmentation}, 
        booktitle={Proceedings of the 2005 IEEE International Conference on Robotics and Automation}, 
        publisher={IEEE}, 
        year={2005}, 
        pages={2277–2282}, 
        ISBN={978-0-7803-8914-4}, 
        DOI={10.1109/ROBOT.2005.1570452},
 }

@book{MOBOT, 
        title={Optimal design of a physical assistive device to support sit-to-stand motions}, 
        ISSN={1050-4729}, 
        DOI={10.1109/ICRA.2015.7140024}, 
        booktitle={2015 IEEE International Conference on Robotics and Automation (ICRA)}, 
        author={Ho Hoang, Khai-Long and Mombaur, Katja D.}, 
        year={2015}, 
        pages={5891–5897} 
 
}

@inproceedings{SMW, address={Zurich}, title={Walking and sit-to-stand support system for elderly and disabled}, ISBN={978-1-4244-9862-8}, url={http://ieeexplore.ieee.org/document/5975365/}, DOI={10.1109/ICORR.2011.5975365}, booktitle={2011 IEEE International Conference on Rehabilitation Robotics}, publisher={IEEE}, author={Hong-Gul Jun and Yoon-Young Chang and Byung-Ju Dan and Byeong-Rim Jo and Byung-Hoon Min and Hyunseok Yang and Won-Kyung Song and Jongbae Kim}, year={2011}, month={Jun}, pages={1–5}, language={en} }

@article{Weight_dis2, title={Weight Discrimination After Anterior Cruciate Ligament Injury: A Pilot Study}, volume={86}, ISSN={00039993}, DOI={10.1016/j.apmr.2004.11.045}, number={7}, journal={Archives of Physical Medicine and Rehabilitation}, author={Héroux, Martin E. and Tremblay, François}, year={2005}, month=jul, pages={1362–1368}, language={en} }

@article{STS_trainer, title={Sit-to-Stand Trainer: An Apparatus for Training “Normal-Like” Sit to Stand Movement}, volume={24}, ISSN={1558-0210}, DOI={10.1109/TNSRE.2015.2442621}, number={6}, journal={IEEE Transactions on Neural Systems and Rehabilitation Engineering}, author={Matjačić, Zlatko and Zadravec, Matjaž and Oblak, Jakob}, year={2016}, month=jun, pages={639–649} }

@inproceedings{Momentum, address={Montreal, Que., Canada}, title={Momentum analysis of sitback failures in sit-to-stand trials}, volume={2}, ISBN={978-0-7803-2475-6}, url={http://ieeexplore.ieee.org/document/579683/}, DOI={10.1109/IEMBS.1995.579683}, booktitle={Proceedings of 17th International Conference of the Engineering in Medicine and Biology Society}, publisher={IEEE}, author={Riley, P.O. and Popat, R. and Krebs, D.E.}, year={1995}, pages={1283–1284}, language={en} }

@article{Joint_angles, title={Trajectory of human movement during sit to stand: a new modeling approach based on movement decomposition and multi-phase cost function}, volume={229}, ISSN={0014-4819, 1432-1106}, DOI={10.1007/s00221-013-3606-1}, number={2}, journal={Experimental Brain Research}, author={Sadeghi, Mohsen and Emadi Andani, Mehran and Bahrami, Fariba and Parnianpour, Mohamad}, year={2013}, month=aug, pages={221–234}, language={en} }

@article{Percentile, title={The measure of man and woman: human factors in design}, author={Tilley, AlVIN R, HENRY DREYFUSS ASSOCIATES}, year={1993}, language={en} }

@article{Failed_STS, title={Biomechanical analysis of failed sit-to-stand}, volume={5}, DOI={10.1109/86.650289}, journal={IEEE transactions on rehabilitation engineering: a publication of the IEEE Engineering in Medicine and Biology Society}, author={Riley, Patrick and Krebs, David and Popat, Rita}, year={1998}, month=jan, pages={353–9} }

@misc{Guldmann, howpublished ={\url{https://www.guldmann.com/ca/products/mobile-lifters/standing-aids/gls5-2-active-lifter}}, abstractNote={The GLS5.2 model is designed to lift the user from a seated to a standing position and is also useful for frequent everyday moves back and forth between bed, chair and bathroom.}, journal={Guldmann}, language={en}, author={Guldmann}, year={2024}, title={GLS5.2 Active lifter}, note = {accessed 29 September 2025}  }

@article{STS_CoM_vel, title={Biomechanics and muscular activity during sit-to-stand transfer}, volume={9}, ISSN={02680033}, DOI={10.1016/0268-0033(94)90004-3}, number={4}, journal={Clinical Biomechanics}, author={Roebroeck, M.E. and Doorenbosch, C.A.M. and Harlaar, J. and Jacobs, R. and Lankhorst, G.J.}, year={1994}, month=jul, pages={235–244}, language={en} }

@article{ZeroG_article, title={ZeroG: Overground gait and balance training system}, volume={48}, ISSN={0748-7711}, DOI={10.1682/JRRD.2010.05.0098}, number={4}, journal={The Journal of Rehabilitation Research and Development}, author={Hidler, Joseph and Brennan, David and Black, Iian and Nichols, Diane and Brady, Kathy and Nef, Tobias}, year={2011}, pages={287}, language={en} }

@article{Feet_placement, title={Muscle activity and balance control during sit-to-stand across symmetric and asymmetric initial foot positions in healthy adults}, volume={71}, ISSN={09666362}, DOI={10.1016/j.gaitpost.2019.04.030}, abstractNote={Background: Rising from a sit to a stand has biomechanical factors that are dependent on initial foot position. Little is known about the eﬀect of initial foot position on leg muscle activation patterns during a sit-to-stand and balance maintenance of stance after a sit-to-stand.}, journal={Gait \& Posture}, author={Jeon, Woohyoung and Jensen, Jody L. and Griffin, Lisa}, year={2019}, month=jun, pages={138–144}, language={en} }

@inproceedings{Platter_pHRI, title={Physical Human-Robot Interaction (pHRI) through Admittance Control of Dynamic Movement Primitives in Sit-to-Stand Assistance Robot}, ISSN={2158-2254}, url={https://ieeexplore.ieee.org/document/9869510/?arnumber=9869510}, DOI={10.1109/HSI55341.2022.9869510}, booktitle={2022 15th International Conference on Human System Interaction (HSI)}, author={Sharma, Bibhu and Pillai, Branesh M. and Suthakorn, Jackrit}, year={2022}, month=jul, pages={1–6} }

@article{STS_review, title={A Review of Intelligent Walking Support Robots: Aiding Sit-to-Stand Transition and Walking}, volume={32}, ISSN={1558-0210}, DOI={10.1109/TNSRE.2024.3379453}, journal={IEEE Transactions on Neural Systems and Rehabilitation Engineering}, author={Sun, Yu and Xiao, Cong and Chen, Lipeng and Chen, Lu and Lu, Haojian and Wang, Yue and Zheng, Yu and Zhang, Zhengyou and Xiong, Rong}, year={2024}, pages={1355–1369} }

@inproceedings{3DOF_Walker, title={A rehabilitation walker with standing and walking assistance}, ISSN={2153-0866}, url={https://ieeexplore.ieee.org/document/4650845/}, DOI={10.1109/IROS.2008.4650845}, booktitle={2008 IEEE/RSJ International Conference on Intelligent Robots and Systems}, author={Chugo, Daisuke and Asawa, Tai and Kitamura, Takuya and Jia, Songmin and Takase, Kunikatsu}, year={2008}, month=sep, pages={260–265} }

@article{Acceleration, title={Study of acceleration of center of mass during sit-to-stand and stand-to-sit in patients with stroke}, volume={28}, ISSN={0915-5287}, DOI={10.1589/jpts.28.2457}, number={9}, journal={Journal of Physical Therapy Science}, author={Na, Eunjin and Hwang, Hyesun and Woo, Youngkeun}, year={2016}, month=sep, pages={2457–2460} }

@article{minimum_muscle, title={The minimum required muscle force for a sit-to-stand task}, volume={45}, ISSN={00219290}, DOI={10.1016/j.jbiomech.2011.11.054}, number={4}, journal={Journal of Biomechanics}, author={Yoshioka, Shinsuke and Nagano, Akinori and Hay, Dean C. and Fukashiro, Senshi}, year={2012}, month=feb, pages={699–705}, language={en} }

@inproceedings{Skywalker, title={Maintaining mobility in older age - design and initial evaluation of the robot SkyWalker for walking \& sit-to-stand assistance}, ISSN={2155-1782}, url={https://ieeexplore.ieee.org/document/9925362/?arnumber=9925362}, DOI={10.1109/BioRob52689.2022.9925362}, booktitle={2022 9th IEEE RAS/EMBS International Conference for Biomedical Robotics and Biomechatronics (BioRob)}, author={Mahdi, Anas and Lin, Jonathan Feng-Shun and Mombaur, Katja}, year={2022}, month=aug, pages={01–08} }

@inproceedings{Force_assistance_system, title={Force Assistance System for Standing-Up Motion}, ISSN={2152-744X}, url={https://ieeexplore.ieee.org/document/4026239/}, DOI={10.1109/ICMA.2006.257779}, booktitle={2006 International Conference on Mechatronics and Automation}, author={Chugo, Daisuke and Kaetsu, Hayato and Miyake, Norihisa and Kawabata, Kuniaki and Asama, Hajime and Kosuge, Kazuhiro}, year={2006}, month=jun, pages={1103–1108} }

@article{LM2S, title={A Dual-Motor Actuator for Ceiling Robots with High Force and High Speed Capabilities}, url={http://arxiv.org/abs/2405.05162}, note={arXiv:2405.05162 [cs, eess]}, number={arXiv:2405.05162}, publisher={arXiv}, author={Lalonde, Ian and Denis, Jeff and Lamy, Mathieu and Girard, Alexandre}, year={2024}, month=may, language={en} }

@article{CMC, title={A new formulation of the coefficient of multiple correlation to assess the similarity of waveforms measured synchronously by different motion analysis protocols}, volume={31}, rights={https://www.elsevier.com/tdm/userlicense/1.0/}, ISSN={09666362}, DOI={10.1016/j.gaitpost.2010.02.009}, number={4}, journal={Gait \& Posture}, author={Ferrari, Alberto and Cutti, Andrea Giovanni and Cappello, Angelo}, year={2010}, month=apr, pages={540–542}, language={en} }

@misc{Guldmann_GPT1, howpublished = {\url{https://www.guldmann.com/ca/products/mobile-lifters/standing-aids/gtp1-transfer-platform}}, author={Guldmann}, language={en}, year={2024}, title={GTP1 Transfer platform}, note = {accessed 26 September 2025} }

@misc{Sara_Stedy, howpublished = {\url{https://www.arjo.com/en-ca/products/patient-handling/standing-and-raising-aid/sara-stedy/}}, author = {Arjo},  abstractNote={Sara Stedy sit-to-stand lift. Enables a single caregiver to assist patients or residents perform sit-to-stand transfers throughout the day.}, language={en}, year={2025}, title={Sara Stedy}, note = {accessed 29 September 2025}  }

@misc{MVN_User_Manual,  title={MVN User Manual}, howpublished = {\url{https://www.movella.com/hubfs/MVN_User_Manual.pdf?__hstc=233546881.1fa5198786ade7bb8cace6bc2dd887f0.1663745187069.1670919674447.1670940876901.93&__hssc=233546881.19.1670940876901&__hsfp=700330257}}, language={en}, note = {Revision 1 April 2023} }

\end{document}